\documentclass[twoside,11pt]{article}
\usepackage{jair, theapa, rawfonts}

\usepackage{url}
\usepackage{caption}
\usepackage{graphicx}
\usepackage{amsmath}
\usepackage{booktabs}
\usepackage{amssymb}
\usepackage{subfigure}
\usepackage{array}
\usepackage{multirow}
\usepackage{color}

\jairheading{70}{2021}{1-27}{12/2020}{04/2021}
\ShortHeadings{RWNE}
{Li, Ji, Peng, He, Song, Zhang, \& Peng}
\firstpageno{1}

\begin{document}

\title{RWNE: A Scalable Random-Walk-Based Network Embedding Framework with Personalized Higher-Order Proximity Preserved}
\renewcommand{\thefootnote}{*}
\author{\name Jianxin Li \email lijx@act.buaa.edu.cn \\
       \name Cheng Ji \email jicheng@act.buaa.edu.cn \\
       \name Hao Peng \email penghao@act.buaa.edu.cn \\
       \name Yu He \email heyu@act.buaa.edu.cn \\
       \addr Beijing Advanced Innovation Center for Big Data and Brain Computing \\
       \addr Beihang University, China
       \AND
       \name Yangqiu Song \email yqsong@cse.ust.hk \\
       \addr Department CSE, HKUST, Hong Kong
       \AND
       \name Xinmiao Zhang \email zhangxinmiao@buaa.edu.cn \\
       \name Fanzhang Peng \email pengfanzhang@buaa.edu.cn \\
       \addr Beihang University, China}

% For research notes, remove the comment character in the line below.
% \researchnote
% \footnotetext{Corresponding Authors: Jianxin Li and Hao Peng.}

\maketitle

\begin{abstract}
Higher-order proximity preserved network embedding has attracted increasing attention.
In particular, due to the superior scalability, random-walk-based network embedding has also been well developed, which could efficiently explore higher-order neighborhoods via multi-hop random walks.
However, despite the success of current random-walk-based methods, most of them are usually not expressive enough to preserve the personalized higher-order proximity and lack a straightforward objective to theoretically articulate what and how network proximity is preserved.
In this paper, to address the above issues,
we present a general scalable random-walk-based network embedding framework,
in which random walk is explicitly incorporated into a sound objective designed theoretically to preserve arbitrary higher-order proximity.
Further, we introduce the random walk with restart process into the framework to naturally and effectively achieve personalized-weighted preservation of proximities of different orders.
We conduct extensive experiments on several real-world networks and demonstrate that our proposed method consistently and substantially outperforms the state-of-the-art network embedding methods.
\end{abstract}

%\vspace{-0.15in}
\section{Introduction}\label{sec:introduction}
Network embedding,
which has recently attracted increasing attention in both academia and industry,
is a general and fundamental technique for representing nodes of the real-word network as vectors in a low-dimensional space while preserving the inherent properties and structures of the network~\shortcite{cui2018survey,goyal2018graph,cai2018comprehensive}.
Such embedding vectors can then be used for a variety of network mining tasks,
such as 
node profiling (classification and clustering)
~\shortcite{sen2008collective,wang2017community}, 
link prediction~\shortcite{ShiZLYYW15,wei2017cross}, 
similarity search~\shortcite{Sun2011a,SunNHYYY12,zhou2017scalable},
etc.
%In the past decade,
%with the development of more scalable machine learning systems and better optimization algorithms, network embedding has been applied to very large-scale networks, which makes it more applicable and worthwhile than ever before.
%Network embedding, especially scalable network embedding for large-scale networks, has recently attracted increasing attention in both academia and industry
%~\shortcite{cui2018survey,goyal2018graph,cai2018comprehensive}.

One basic requirement of network embedding is that the learned vectors of nodes should preserve the network structures. % and its inherent properties~\shortcite{wang2017community}.
Along with this direction, many network embedding methods are proposed to preserve the first-order proximity which expresses the local pairwise structure indicated by the observed edges between nodes
(e.g., \shortciteR{roweis2000nonlinear,belkin2002laplacian,SpectralClustering2011,GF2013}), %YangSLT17
or further to preserve the second-order proximity between a pair of nodes which implies the similarity between their neighborhood structures (e.g., ~\shortciteR{LINE2015,SDNE2016}).

Despite their success, in recent years, % 
more and more works~\shortcite{GraRep2015,HOPE2016,YangSLT17,zhang2018arbitrary} have demonstrated that, 
besides the first- and second- order proximity directly indicated by pairwise edges,
the higher-order proximity is also one of tremendous importance in capturing the underlying structures of the network. % and learning the useful embedding vectors. 
%and thus can provide much valuable information for learning the embedding vectors.
\begin{itemize}
    \item \textbf{Different-order proximities describe the network structures from different levels of scope, which give us much valuable information with different granularities.} Thus, embeddings with the lower-order proximity alone or even any certain-order proximity do not necessarily perform best on all networks and target applications~\shortcite{perozzi2017don,zhang2018arbitrary}. For example, in classification tasks with coarse-grained classes, the higher-order proximity is likely to be more helpful than lower-order proximity. 
    \item \textbf{Real-world networks are usually very sparse, with only a small number of edges observed.} That is, the observed first-order proximity and even second-order proximity may not be sufficient to reflect the underlying relations between nodes~\shortcite{LINE2015}. Therefore, to address the network sparsity issue, it is also very important to incorporate higher-order proximity to capture more available information.
\end{itemize}

%\vspace{-0.05in}
Although many works have been proposed to preserve the higher-order proximity in network embedding, %and thus generate high-quality embedding vectors,
most of them are developed to explicitly exploit the higher-order proximity matrix by the technique of matrix factorization (e.g., \shortciteR{GraRep2015,HOPE2016,YangSLT17}) or deep learning (e.g., \shortciteR{SDNE2016,DNGR2016}), which are known to have scalability issues when dealing with large-scale networks~\shortcite{YangSLT17}.
%In principle, an accurate computation of higher-order proximity matrix is usually time and space consuming with $O(|\!V\!|^2\!)$ complexity and thus not feasible for large-scale networks~\shortcite{YangSLT17}.

% \textcolor{blue}{
To be more efficient, inspired by the Skip-gram algorithm % in natural language processing~
\shortcite{Mikolov2013},
random-walk-based network embedding algorithms have also been well developed~\shortcite{DeepWalk2014,LINE2015,Node2Vec2016,perozzi2017don,he2019hetespaceywalk}. %which are known for having superior scalability for large-scale networks and exploring higher-order neighborhood via multi-hop random walks.
% In general, these algorithms use a two-step approach to generate node embeddings:
% random walk to produce the node sequences and Skip-gram~\shortcite{Mikolov2013} to learn vectors.
%They first perform random walks on a network to generate nodes sequences and then run the Skipgram algorithm over these sequences to generate node embeddings.
Although these random-walk-based algorithms are known for having superior scalability for large-scale networks and having the ability for exploring higher-order neighborhood via multi-hop random walks, there are still some issues:
% }
\begin{itemize}
    \item They either treat different-order neighborhood equivalently~\shortcite{DeepWalk2014} which may
not be expressive enough to incorporate a personalized combination of proximities of different orders, 
or use a somewhat complex second-order biased random walk~\shortcite{Node2Vec2016} which can be costly when pre-computing its third-order transition probability hypermatrix~\shortcite{zhang2018arbitrary}.
    \item In essence, these algorithms convert the network embedding problem as the word embedding problem by treating a node as a word and pre-generating the node ``corpus'' via random walks. 
{As a consequence, %these algorithms are inferior because 
they have no specific objective to articulate what and how network proximity is preserved and have no sound theory to estimate the essential role of random walk playing in network embedding}, which also limits their superiority. % desirability.
% making them inadequate and fragile in theory. 
%  unexplainable and unsubstantial in theory. %fragile inferior
\end{itemize}

In this paper, to address the above issues, 
we present a general scalable \textbf{\emph{R}}andom-\textbf{\emph{W}}alk-based \textbf{\emph{N}}etwork \textbf{\emph{E}}mbedding framework (called \textbf{\emph{RWNE}}),
in which 
%both the first-order and higher-order proximities 
arbitrary higher-order proximity 
of the network can be explicitly preserved with a sound objective carefully designed to simultaneously capture both the local pairwise similarity and the global listwise equivalence between nodes.
%and the effectiveness and efficiency are simultaneously guaranteed. % due to the random walk sampling optimization.
%to fully capture the network structures,
%we model each-order proximity in terms of both the direct strength and the relative differentiation, which characterize the local similarity and global equivalence between nodes respectively.
%In practice,  for arbitrary higher-order proximity,
%we carefully design two sound objectives, which characterize the direct similarity and relative equivalence between nodes respectively.
%Moreover, systematically, theoretically
More than that, to make the framework efficient and practical for large-scale networks,
we theoretically show that the above objective can be equivalently optimized by sampling the nodes via the random walk process with the probability proportional to the proximity,
which  clarifies 
why and how we can use random walks to preserve arbitrary user-specified network proximity and reversely explains what and how network proximity is preserved for an arbitrary user-specified random walk.
%We further propose to leverage 
Further, we introduce 
the random walk with restart process to naturally and effectively achieve  personalized-weighted preservation of different-order proximities 
%which can naturally adjust the prestiges of different proximities %orders 
with an elegant attenuation function controlled by a personalized teleport probability.
Finally,
we conduct extensive experiments on six real-world networks %of a wide range of sizes %in  terms of both the node classification and node clustering tasks.
over three classical network mining tasks: multi-label node classification, node clustering, and link reconstruction.
The experimental results demonstrate that our proposed method consistently and substantially outperforms the state-of-the-art network embedding methods.
%extensive experiments on several real-world networks demonstrate that our proposed method consistently and significantly outperforms the state-of-the-art network embedding methods 
%in several applications of network embedding, including network reconstruction,
%link prediction
To summarize, the main contributions of our work are as follows:
\begin{enumerate}
	\vspace{-0.04in}
\item[1.]
We systematically present a general scalable random-walk-based network embedding framework \textbf{\emph{RWNE}}\footnote{The datasets and code are released at \url{https://github.com/RingBDStack/RWNE}.}, 
%to effectively and efficiently preserve arbitrary higher-order proximity,
in which random walk is efficiently and explicitly incorporated into a sound objective designed theoretically to preserve arbitrary higher-order proximity.
%in which a sound objective is explicitly designed to effectively preserve arbitrary higher-order proximity and then is efficiently optimizated by random-walk sampling. % with a sound theory.
\vspace{-0.04in}
\item[2.]
We further introduce the random walk with restart process to practically preserve the personalized higher-order proximity which naturally weights different-order proximities with an elegant attenuation function controlled by a personalized teleport probability.
\vspace{-0.04in}
\item[3.]
%We conduct extensive experiments on six real-world networks of a wide range of sizes.
%The empirical results demonstrate that our model significantly outperforms the current state-of-the-art network embedding methods.
We conduct extensive experiments on several real-world networks and demonstrate that our proposed model consistently and considerably outperforms the state-of-the-art network embedding methods.

\end{enumerate}

%The effectiveness of the embedded vectors is evaluated by experiments on three large-scale real-world networks.
%The results 
%prove %demonstrate 
%that HPNE significantly outperforms the current state-of-the-art network embedding methods.

%\vspace{-0.15in}
\section{Related Work}\label{sec:related_work}
%\vspace{-0.05in}
%Network embedding is a general and fundamental problem to represent nodes of a network as vectors in a low-dimensional vector space.
%Such a low-dimensional embedding is very useful in a variety of applications,
%such as 
%node profiling (classification or clustering)
%~\shortcite{sen2008collective,bhagat2011node,wang2017community},
%link prediction~\shortcite{ShiZLYYW15,wei2017cross}, 
%similarity search~\shortcite{Sun2011a,zhou2017scalable}, etc.
Network embedding has aroused lots of research interest for a long time~\shortcite{cui2018survey,cai2018comprehensive,wang2020brief}.
The earlier network embedding algorithms, also called graph embedding, are studied as a dimension reduction problem,
such as LLE~\shortcite{roweis2000nonlinear}, Laplacian eigenmaps~\shortcite{belkin2003laplacian}, etc.
%However,
These methods focus on the first-order proximity 
%characterized by the (normalized) adjacency matrix or Laplacian matrix.
which captures the local structure information of the network.

\textbf{Matrix factorization.}
Recently,
to sufficiently explore the network structure from different levels of scope,
a bunch of methods has been proposed to preserve the higher-order proximity in network embedding.
Most of these methods are proposed to explicitly factorize a higher-order proximity matrix,
%by applying decomposition techniques (e.g., SVD), 
such as 
GraRep~\shortcite{GraRep2015}, 
HOPE~\shortcite{HOPE2016}, 
M-NMF~\shortcite{wang2017community},
%NEU~\shortcite{YangSLT17},
NetMF~\shortcite{qiu2018network},
AROPE~\shortcite{zhang2018arbitrary},
etc.
However, in principle,
as the computation and storage of higher-order proximity matrices are generally at least $O(|V|^2)$ complexity,
these matrix-factorization methods often have efficiency issues when dealing with large-scale networks. 

\textbf{Deep Learning.}
Besides,
deep learning is also studied in preserving higher-order proximity. For example,
SDNE~\shortcite{SDNE2016} first applies a deep auto-encoder to preserve both $1$st- and $2$nd-order proximity.
DNGR~\shortcite{DNGR2016} further uses a stacked denoising auto-encoder to preserve higher-order proximity.
Unfortunately, same as matrix-factorization methods, these methods also confront efficiency issues.
Specially,
deep convolution networks are popularly applied on graphs in very recent years (e.g.,~\shortciteR{GCN2016,GAN2017,peng2019hierarchical}),
which are studied as supervised/semi-supervised feature learning models with node attributes/features %and node labels 
incorporated.
We also compare with these methods, but it is noteworthy that in this paper we focus on the most fundamental case that only the network structure information is available.

\textbf{Random walk.}
On the other hand,
due to the superior scalability,
%inspired by the Skip-gram algorithm~\shortcite{Mikolov2013},
%word embedding~\shortcite{Mikolov2013}, 
random-walk-based network embedding algorithms have also been well developed.
%which are known to have superior scalability for large-scale networks and
%adaptability for %handlability %adaptability for %explore
%higher-order neighborhood via long-distance random walks.
% \textcolor{blue}{
A two-step framework is applied for generating the node embeddings in these methods.
First, they perform random walks on a network to generate node sequences.
Then they run the Skip-gram algorithm over these sequences to generate node embeddings.
% first perform random walks on a network to generate nodes sequences and then run the Skipgram algorithm~\shortcite{Mikolov2013} over these sequences to generate node embeddings.
For example,
DeepWalk~\shortcite{DeepWalk2014} uses %truncated random walks 
uniform random walks 
to generate node sequences and then runs the Skip-gram algorithm. %over these sequences 
%to generate embeddings. 
The major drawback of DeepWalk is that it treats different-order neighborhoods equivalently and thus maybe not expressive enough to incorporate a personalized combination of proximities of different orders.
node2vec~\shortcite{Node2Vec2016} further generalizes a second-order %Markovian process named
biased random walk
% for more flexibility, which aims 
to seek a trade-off  between breadth-first %(BFS) 
and depth-first %(DFS) 
graph searches. % and hence produces more informative embeddings.
Although node2vec can incorporate a biased combination of different-order proximities, it is usually costly for computing the second-order transition probability hypermatrix of the proposed second-order random walk.
% }
% which can be costly when pre-computing its second-order transition probability hypermatrix
% Moreover,
% these existing random-walk based algorithms actually convert the network embedding problem into the word embedding problem by treating a node as a word and pre-generating the node ``corpus'' via random walks. 
% As a consequence, % it is known that
% they generally lack sound objectives to articulate what and how network proximity is preserved.
Moreover,
despite their success,
as introduced in Section \ref{sec:introduction}, there are still some issues.
In this paper, instead of the above two-step framework,
we focus on a straightforward framework 
%by explicitly leveraging random walk to preserve higher-order proximity.
by explicitly incorporating random walk into a sound objective designed theoretically to preserve personalized higher-order proximity.

\vspace{-0.05in}
\section{Definitions and Preliminaries}\label{sec:preliminaries}
In this section, we first introduce the problem of network embedding, and then formally define the measures of higher-order proximity to characterize network structures. 
% In this section, we first introduce the basic concepts about network embedding, and then present a formal definition of high-order proximity.
We first formalize a network as follows.

{\it \textbf{Network}.
A network is defined as a directed graph 
${\mathnormal G}\!\!=\!\!({\mathnormal V}, {\mathnormal E})$, 
where ${\mathnormal V}\!\!=\!\!\{v_1,v_2,\cdots\!,v_n\}$ is the set of nodes and $n$ is its size,
${\mathnormal E}\!\!=\!\!\{e_{i,j}\}_{i,j\!=\!1}^n$ is the set of edges. 
Each edge $e_{i,j}$ is equally as a linked node pair $(v_i,v_j)$ and is associated with a weight $w_{i,j}\!\geqslant \!0$, which indicates the strength of the edge.
If there is no observed edge from $v_i$ to $v_j$, then $w_{i,j}\!=\!0$.
Specially, for an undirected graph, we have $w_{i,j}\!\equiv\!w_{j,i}$.
For an unweighted graph, we have $w_{i,j}\!\in\! \{0,1\}$.
}

% % Given a network, without loss of generality, we assume it is directed (an undirected edge can be considered as two directed edges with opposite directions and equal weights).
% % Note that given a network,
% In a real network,
% the edges can be %either
% directed or undirected, 
% and the weights can be %either 
% binary or take any real value.
% Without loss of generality, we assume edges are always directed and 
% %non-negatively 
% weighted. 
% If an edge $e_{i,j}$ is undirected, we have $(v_i,v_j)\!\!\equiv\!\!(v_j,v_i)$ and $w_{i,j}\!\equiv\!w_{j,i}$. 
% If a weight $w_{i,j}$ is binary, we have $w_{i,j}\!\in\! \{0,1\}$. 
% And while negative weights are possible, in this study we only consider non-negative weights.

% Following the above definition,
% the adjacency matrix of an information network is defined as 
% $\widetilde{A}\in \mathbb{R}^{|V|\times|V|}$, 

We further denote $A\!\in\! \mathbb{R}^{n\!\times\!n}$ as the adjacency matrix of the network, where each entry 
$A_{i,j}\!\!=\!\!w_{i,j}$ represents the weight of the edge from $v_i$ to $v_j$; 
denote $D\!\in\! \mathbb{R}^{n\!\times\! n}$ as the weight matrix where each diagonal entry $D_{i,i}\!=\!\sum_{j\!=\!1}^{n}\!A_{i,j}$
represents the summed ``out-weight'' of $v_i$ and other entries are zero.
Then, 
$\Lambda\!=\!D^{-\!1}\!A$ 
is the normalized adjacency matrix, i.e., the one-step transition probability matrix for random walks on the network, where the sum of each row is equal to one.
% Similarly, we also have the Laplacian matrix $L\!=\!D\!-\!A$ and the normalized Laplacian matrix $\Gamma\!=\!D^{-1}\!L\!=\!I\!-\!\Lambda$.

\begin{figure}[t]
% 	\small
%		\setlength{\abovecaptionskip}{-0.01cm}
% 		\setlength{\belowcaptionskip}{-0.1cm}
	\centering
	\includegraphics[width=0.5\textwidth]{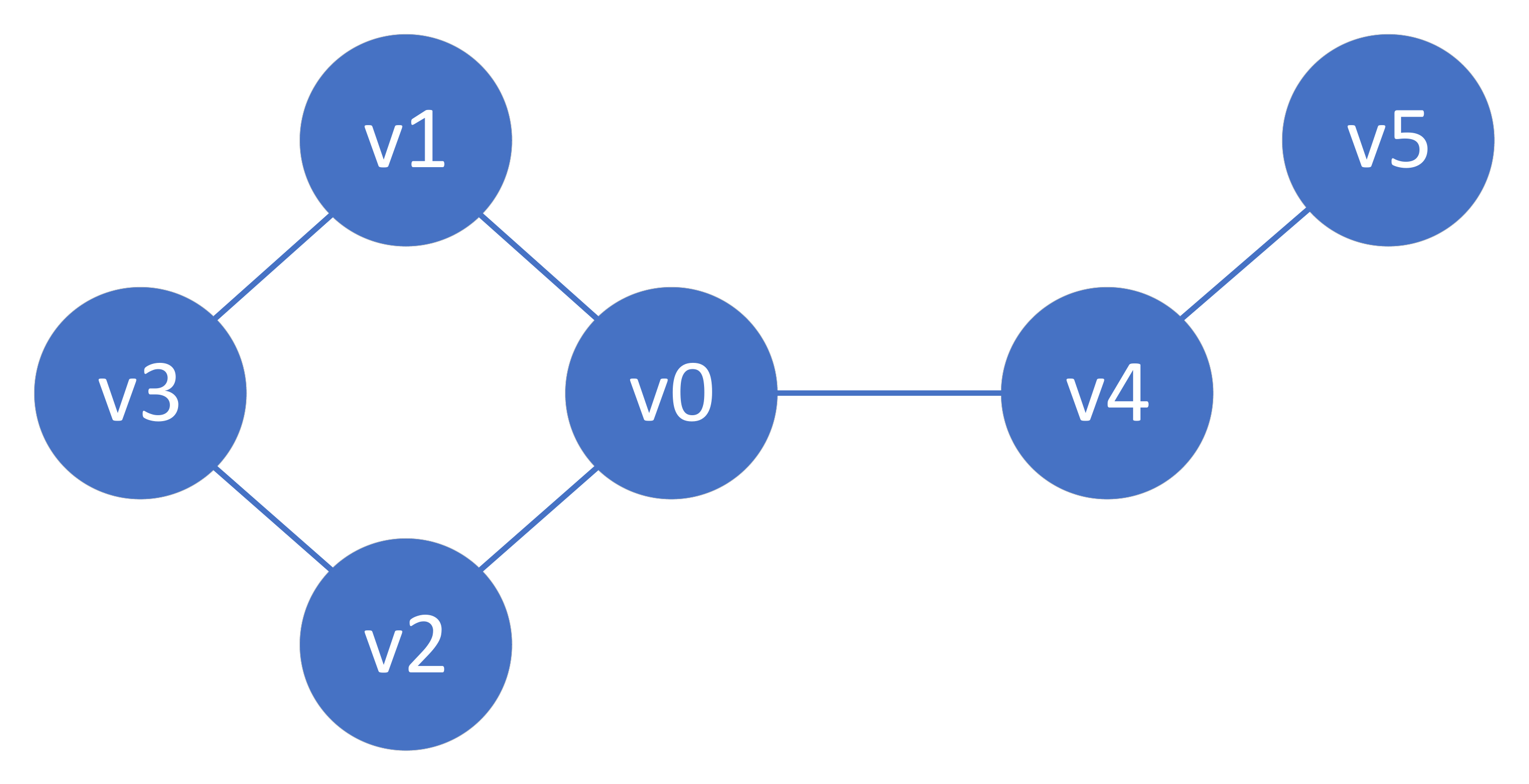}
	\caption{An illustrative network $G_e$ with six nodes.}
	\label{Fig:example_network}
% 	\vspace{-0.1in}
\end{figure}

\begin{example}\label{Exam:HIN}
Take a simple graph ${\mathnormal G_e}\!\!=\!\!({\mathnormal V}, {\mathnormal E})$ shown in Figure~\ref{Fig:example_network} as an example, with $w_{i,j}\!\equiv\!w_{j,i}$=1. As defined above, we have:

${\mathcal V} = \{v_0, v_1, v_2, v_3, v_4, v_5\}$,

${\mathcal E} = \{e_{0,1}, e_{0,2}, e_{0,4}, e_{1,0}, e_{1,3}, e_{2,0}, e_{2,3}, e_{3,1}, e_{3,2}, e_{4,0}, e_{4,5}, e_{5,4}\}$.

The adjacency matrix and weight matrix are
$$
A =
\begin{bmatrix}
0 & 1 & 1 & 0 & 1 & 0 \\
1 & 0 & 0 & 1 & 0 & 0 \\
1 & 0 & 0 & 1 & 0 & 0 \\
0 & 1 & 1 & 0 & 0 & 0 \\
1 & 0 & 0 & 0 & 0 & 1 \\
0 & 0 & 0 & 0 & 1 & 0 \\
\end{bmatrix}
, \quad
D =
\begin{bmatrix}
3 & 0 & 0 & 0 & 0 & 0 \\
0 & 2 & 0 & 0 & 0 & 0 \\
0 & 0 & 2 & 0 & 0 & 0 \\
0 & 0 & 0 & 2 & 0 & 0 \\
0 & 0 & 0 & 0 & 2 & 0 \\
0 & 0 & 0 & 0 & 0 & 1 \\
\end{bmatrix}
.
$$

Thus, the normalized adjacency matrix can be calculated as
$$
\Lambda = D^{-1}A =
% \begin{bmatrix}
% 3 & 0 & 0 & 0 & 0 & 0 \\
% 0 & 2 & 0 & 0 & 0 & 0 \\
% 0 & 0 & 2 & 0 & 0 & 0 \\
% 0 & 0 & 0 & 2 & 0 & 0 \\
% 0 & 0 & 0 & 0 & 2 & 0 \\
% 0 & 0 & 0 & 0 & 0 & 1 \\
% \end{bmatrix}
% \cdot
% \begin{bmatrix}
% 0 & 1 & 1 & 0 & 1 & 0 \\
% 1 & 0 & 0 & 1 & 0 & 0 \\
% 1 & 0 & 0 & 1 & 0 & 0 \\
% 0 & 1 & 1 & 0 & 0 & 0 \\
% 1 & 0 & 0 & 0 & 0 & 1 \\
% 0 & 0 & 0 & 0 & 1 & 0 \\
% \end{bmatrix}
% =
\begin{bmatrix}
0 & 1/3 & 1/3 & 0 & 1/3 & 0 \\
1/2 & 0 & 0 & 1/2 & 0 & 0 \\
1/2 & 0 & 0 & 1/2 & 0 & 0 \\
0 & 1/2 & 1/2 & 0 & 0 & 0 \\
1/2 & 0 & 0 & 0 & 0 & 1/2 \\
0 & 0 & 0 & 0 & 1 & 0 \\
\end{bmatrix}
.
$$
\end{example}

{\it \textbf{Network Embedding}.
Given a network ${\mathnormal G} \!\!=\!\! ({\mathnormal V}, {\mathnormal E})$, the problem of network embedding aims to embed the network into a low-dimensional space while preserving the network structures, i.e., learn a mapping function $f_{\small{G}}\!:\!v_i\!\in\!\!V\!\!\to\!\!r_i\!\!\in\!\!\mathbb{R}^d\!$, % where $r_i$ is the embedding representation vector of $v_i$
where $d\!\!\ll\!\! |V|$ and $f_G$ preserves the relations between nodes.
}

As defined above, to conduct the embedding, the network structures, i.e., the relations between nodes, must be preserved as much as possible.
In practice, we adopt the following proximity measures to quantify the network structure information to be preserved in the embedded space.
We first define the first-order proximity as follows.

{\it \textbf{First-order Proximity}.
The first-order proximity is defined to measure the adjacent structures of a network.
For each pair of nodes $(v_i,v_j)$, if they are linked by an edge, there exists positive first-order proximity between them, and the weight $w_{i,j}$ on that edge indicates the strength of the proximity.
If no edge is observed between $v_i$ and $v_j$, their first-order proximity is zero.
Further, for all pairs of nodes, there is naturally a first-order proximity matrix indicated by the adjacency matrix of the network.
}

In practice, since the weights of edges in a network can diverge over a very wide range, we use the normalized adjacency matrix (i.e., the one-step transition probability matrix) $\Lambda$ as the formal first-order proximity matrix, where each normalized entry $\Lambda_{i,j}$ is the first-order proximity of node pair $(v_i,v_j)$, which also represents the transition probability of one-step random walk from $v_i$ to $v_j$.
% each normalized vector 
% $\Lambda_i\!=\!(\Lambda_{i,1},\Lambda_{i,2},\cdots\!,\Lambda_{i,|\!V\!|})$ 
% is the first-order proximity vector of $v_i$ with all the other nodes which is also the probability distribution of one-step random walk from $v_i$ to other nodes.
% \textcolor{blue}{
It is necessary to consider the structural characteristics of the network from local and global perspectives, which has been proved by many researches, such as feature selection \shortcite{liu2013global}, semi-supervised classification \shortcite{kang2019robust,kang2021structured}, clustering ~\shortcite{ren2020simultaneous,kang2020relation}, etc. The defined first-order proximity can measure the adjacent structures in both local and global aspects:
% }
\begin{itemize}
\item In local aspect, 
the first-order proximity implies that two nodes %in real-world networks 
are similar if they are linked by an observed edge.
For example, 
bloggers following each other in a social network tend to share similar interests;
papers citing to each other in a citation network tend to talk about similar topics.

\item In global aspect,
the relation between two nodes is also determined by their common neighbors.
% global equivalence with the entire node set of the network. 
For example, people sharing many common friends in a social network are likely to share similar interest and become friends.
That is, even if two nodes are not directly connected, we can capture their relation through their neighbors.
\end{itemize}

Specially, we redefine such local similarity between each two nodes $(v_i,v_j)$ directly determined by their first-order proximity entry $\Lambda_{i,j}$ as the first-order local proximity.
By preserving the first-order local proximity, we are able to characterize the local adjacent structure of the network.

For further convenience, we here define two vectors: $\Lambda_i\!=\!(\Lambda_{i,1},\Lambda_{i,2},\cdots\!,\Lambda_{i,|V|})$,
and $\Lambda_{\cdot,i}\!=\!(\Lambda_{1,i},\Lambda_{2,i},\cdots\!,\Lambda_{|V|,i})$.
More formally,
for three nodes $v_i,v_{j_1},v_{j_2}$, if their first-order proximities satisfy $\Lambda_{i,j_1}\!\!=\!\!\Lambda_{i,j_2}$, 
i.e., 
$v_i$ randomly walks to  $v_{j_1}$ and  $v_{j_2}$  with the same probability, 
then $v_{j_1}$ and $v_{j_2}$ share the equivalent role for %in view of 
$v_i$.
Further, if %in view of 
for each one of the entire node set, $v_{j_1}$ and $v_{j_2}$ always share the equivalent role, i.e., $\Lambda_{\cdot,j_1}\!\!\equiv\!\!\Lambda_{\cdot,j_2}$, 
then $v_{j_1}$ and $v_{j_2}$ have an equivalent global structure role in the network.
For example, 
if there are such two papers that all the papers in the citation network who cite one of them will also cite the other, the two papers are very probably to talk about the same topic and have the equivalent significance in academia.
Therefore, we can capture the global adjacent structure by preserving such global equivalence between nodes.
Specially, we refer to such global equivalence between each two nodes $(v_{j_1},v_{j_2})$ determined by the similarity of the two vectors $\Lambda_{\cdot,j_1}$ and $\Lambda_{\cdot,j_2}$ as the first-order global proximity.

\begin{example}
For the node $v_1$ and node $v_2$ in $G_e$, although the two nodes are not directly connected, they should have stronger similarities because they have multiple common neighbors node $v_0$ and node $v_3$, which can be judged by the first-order global proximity:
$$
\Lambda_{\cdot,v_1} = \Lambda_{\cdot,v_2} =
\begin{bmatrix}
1/3 & 0 & 0 & 1/2 & 0 & 0 \\
\end{bmatrix}
.
$$
\end{example}

Intuitively, it is necessary for network embedding to preserve both the
first-order local proximity and the first-order global proximity. The two proximities characterize the adjacent structures of the network in local and global aspects, respectively. 
% As a result, both the local adjacent structure and global adjacent structure in the network can be well embedded.
% Note that each offdiagonal nonzero entry of the first-order proximity matrix corresponds to an edge in the network. However, real-world networks are always very sparse, with a very small proportion edges. 
However, there are more non-adjacent structures in the network that cannot be described by first-order proximity.
For example, two nodes can be similar even if they neither have a local edge nor a common neighbor.
Actually, in many real-world networks, the adjacent structures observed are only a small proportion, with many others missing~\shortcite{LINE2015}.
That is, the first-order proximity matrix is usually sparse and thus is not sufficient to capture network structures.
To address the sparsity, the higher-order proximity must be preserved.
In fact, even if two nodes have no edge or neighbor, they can also be related if there is a path (i.e., a sequence of directed edges) between them, which could be regarded as a long-distance ``edge''.
Therefore, we define the higher-order proximity as follows.

{\it \textbf{Higher-Order Proximity}.
The higher-order proximity is defined to measure the long-distance structures of a network. %i.e., the non-adjacent relations between nodes linked by paths in the network.
For each pair of nodes $(v_i,v_j)$, if they are linked by a path, i.e., 
we can walk from $v_i$ to $v_j$ along several edges, 
% there exists positive higher-order proximity between them. 
they have a higher-order proximity.
% And the more such paths, the stronger their higher-order proximity.
Specially, if the length of the path is $k$, they have a $k$-th order proximity; if there is no $k$-length path between them, their $k$-th order proximity is $0$.
}

Similar to the first-order proximity which represents the one-step transition probability, we can use the probability of walking from $v_i$ to $v_j$ along paths to represent the strength of the higher-order proximity.
Specially, the $k$-th order proximity matrix can be computed as:
% the $k$-power multiplication of first-order proximity matrix: 
% $\Lambda^k=\Lambda\cdot \Lambda\cdots \Lambda,\ k\!=\!2,3,\cdots$, 

% \setlength\abovedisplayskip{1pt}
% \setlength\belowdisplayskip{1pt}
\begin{equation}
\label{Eq:kth_proximity_matrix}
\Lambda^k=\underbrace{\Lambda\cdot \Lambda\dots \Lambda}_k, \ \ k=2,3,\cdots,
\end{equation}
where each entry $\Lambda_{i,j}^k$ is the $k$-th order proximity of node pair $(v_i,v_j)$, which also represents the $k$-step transition probability from $v_i$ to $v_j$.
Also similar to the first-order proximity,
the $k$-th order proximity can be further derived as $k$-th order local proximity and $k$-th order global proximity to characterize  ``$k$-th order'' network structures (i.e., the non-adjacent relations between nodes linked by $k$-length paths) in local and global aspects, respectively.
For each pair of nodes $(v_i,v_j)$, 
the $k$-th order local proximity represents their local similarity directly determined by the entry $\Lambda_{i,j}^k$,
and the $k$-th order global proximity represents their global equivalence determined by the similarity of $\Lambda_{\cdot,i}^k$ and $\Lambda_{\cdot,j}^k$.

\begin{example}
The higher-order proximity matrices of $G_e$ are given here when $k$ equals to 3 and 5, with the case when k equals to 1 shown in Example~\ref{Exam:HIN}:
$$
\Lambda^{3} =
\begin{bmatrix}
0 & 1/3 & 1/3 & 0 & 1/3 & 0 \\
1/2 & 0 & 0 & 5/12 & 0 & 1/12 \\
1/2 & 0 & 0 & 5/12 & 0 & 1/12 \\
0 & 5/12 & 5/12 & 0 & 1/6 & 0 \\
1/2 & 0 & 0 & 1/6 & 0 & 1/3 \\
0 & 1/6 & 1/6 & 0 & 2/3 & 0 \\
\end{bmatrix}
,
\Lambda^{5} =
\begin{bmatrix}
0 & 1/3 & 1/3 & 0 & 1/3 & 0 \\
1/2 & 0 & 0 & 3/8 & 0 & 1/8 \\
1/2 & 0 & 0 & 3/8 & 0 & 1/8 \\
0 & 3/8 & 3/8 & 0 & 1/4 & 0 \\
1/2 & 0 & 0 & 1/4 & 0 & 1/4 \\
0 & 1/4 & 1/4 & 0 & 1/2 & 0 \\
\end{bmatrix}.
$$
For instance, it is directly observed that the node $v_1$ and node $v_5$ are related under the view of 3-rd order proximity and 5-th order proximity, which cannot be captured by the first-order proximity (adjacency matrix) for the two nodes are not actually connected in the graph. Moreover, it can be noticed that the relation is different under $\Lambda^{3}$ and $\Lambda^{5}$, which indicates that different higher-order proximity holds different structural semantics.
\end{example}

Note that first-order proximity is the special case of $k$-th order proximity when $k\!=\!1$. 
% If not specified, we denote the high-order proximities as the combination of $1$st-, $2$nd-, $\cdots$, and $k$-th order proximities.
% \textcolor{blue}{
In this paper, without loss of generality, we use ``the higher-order proximity'' to represent an unspecified $k$-th order proximity with $k\!>\!1$, and use ``high-order proximities'' to represent the combination of $1$st-, $2$nd-, $\cdots$, and $k$-th order proximities.
% }
By preserving these high-order proximities, we are able to sufficiently characterize both adjacent and long-distance network structures in both local and global aspects, which is very effective in solving the sparsity problem and achieving high-quality embeddings.

% By combining the local similarities and global equivalences between nodes based on these high-order proximities, both the local and global, adjacent and non-adjacent structures of the network could be fully modeled.
% % which is especially significant for sparse networks.
% Next, we will present such a novel method to efficiently and scalably model network structures and get high-quality embeddings.

% We can model non-adjacent structures by using high-order proximity, which is especially significant for sparse networks.
% However, many existing graph embedding algorithms only preserve the local adjacent structures by using first-order proximity.
% Next, we introduce a novel method to utilize the high-order proximities.

%\vspace{-0.05in}
\section{The RWNE Framework}\label{sec:method}
In this section, we present a general scalable embedding framework \emph{RWNE} for network embedding and further introduce the optimization method with random walk simulation.
%\vspace{-0.02in}
\subsection{Problem Formulation}
%\vspace{-0.02in}
%Given a network ${\mathnormal G}\!\!=\!\!({\mathnormal V}, {\mathnormal E})$,
We formulate the normalized adjacency matrix (denoted as $A$ in this section) as the first-order proximity matrix, which captures the direct
neighbor relations between nodes~\shortcite{LINE2015,YangSLT17}.
Specially, such first-order proximity can be alternatively viewed as the transition probability of a single step of the random walk over the network.
Then,
in the probabilistic setting based on random walk, we can easily generalize it to $k$-th order proximity $A^k$~\shortcite{GraRep2015}:
the transition probability of a random walk %starts from $v_i$ and walks to $v_j$ 
with exactly $k$ steps,  which represents the %strength of 
$k$-hop relations between nodes.
As proximities of different orders explore the relations from different levels of scope, which all can provide valuable information to guide the embedding,
a desirable embedding model for real-world networks must be capable of preserving a delicate integrated higher-order proximity %matrix $D$
 which combining the proximities of different orders as follows:

% 	\setlength\abovedisplayskip{2.pt}
% 	\setlength\belowdisplayskip{2.pt}
% 	\begin{small}
	\begin{equation}
	\label{Eq:higher_order_proximities}
	D = \beta_1 A + \beta_2 A^2 + \cdots + \beta_k A^k, \ \ \ \ k = 1,2,\cdots,\infty,
	\end{equation}
% 	\end{small}
where $\beta_k$ is the weight to control the prestige of $k$-th order proximity $A^k$, 
and %usually 
the sum of all weights $\sum_{i=1}^{k}\beta_i \!\!=\!\! 1$.
%The integrated higher-order proximity $D$ is comprehensive and superior proximity measure than a single order proximity.
%Therefore, there are generally two problems: how to preserve the higher-order proximity and how to determine appropriate weights to incorporate proximities of different orders.

%In the rest of this section, incorporate

\begin{example}
In the case where $k=5$ and $\beta_1=\dots=\beta_5=0.2$, the integrated higher-order proximity matrix of $G_e$ is
$$
D =  
\begin{bmatrix}
0.2 & 0.2 & 0.2 & 0.13 & 0.2 & 0.07 \\
0.3 & 0.16 & 0.16 & 0.26 & 0.08 & 0.04 \\
0.3 & 0.16 & 0.16 & 0.26 & 0.08 & 0.04 \\
0.2 & 0.26 & 0.26 & 0.18 & 0.08 & 0.02 \\
0.3 & 0.08 & 0.08 & 0.08 & 0.23 & 0.22 \\
0.2 & 0.08 & 0.08 & 0.03 & 0.43 & 0.17 \\
\end{bmatrix}
.
$$
\end{example}

In the rest of this section,
%we focus on a general random walk based network embedding framework (named {\it RWNE}), which is able effectively and efficiently
we will systematically present a general scalable random-walk-based network embedding framework, called {\it RWNE}, which is able to effectively and efficiently preserve the above integrated higher-order proximity.
Without loss of generality,
we first describe the {\it RWNE} model to directly preserve a general form of higher-order proximity $D$ with arbitrary weights. %without any optimization,
%in which we carefully design a sound objective to explicitly capture both the local pairwise similarity and global listwise equivalence between nodes (the framework is shown in Figure \ref{Fig:framework}).
Then, we show a superior optimization with random-walk simulation, which makes the model computationally efficient and scalable for large-scale networks.
%Specially,
We further introduce the random walk with restart process to naturally adjust the prestige of different-order proximities by a personalized teleport probability.
%simply but effectively achieve a personalized-weighted preservation of different-order proximities 
%which can naturally adjust the prestiges of different proximities %orders 
%practically preserve the personalized higher-order proximity 
%which naturally weight different-order proximities with an elegant attenuation function 
%controlled by a personalized teleport probability.
% rwr can naturally weight different-order proximities with a decreasing function, which simply but effectively solves the combination problem of different-order proximities.
Finally, we briefly analyze the time complexity of {\it RWNE}.
%we briefly analyze the complexity of the model.

%\vspace{-0.05in}
%pairwise and listwise 
\subsection{Preserving Higher-Order Proximity}
The higher-order proximity is divided into two parts in our work: the local pairwise similarity and the global listwise equivalence. To preserve both two aspects, we further design a joint objective function for our model.
%\vspace{-0.02in}
\label{local_and_global}
\subsubsection{Local Pairwise Similarity}
Given a higher-order proximity matrix $D$ integrating all orders from the $1$-st to the $k$-th as defined in Eq. (\ref{Eq:higher_order_proximities}),
it is straightforward that each entry $D_{ij}$ implies the local pairwise similarity between each pair of nodes $(v_i,v_j)$ in view of the integrated relation from different levels of scope (from $1$-hop to $k$-hop).
If $D_{ij}\!>\!0$, there is a similarity between $v_i$ and $v_j$, and the larger $D_{ij}$ is, the more similar $v_i$ and $v_j$ are;
If $D_{ij}\!=\!0$, they have no similarity.
Therefore,
we can directly use each entry $D_{ij}$ to constrain the similarity of the embedding vectors of each pair of nodes $(v_i,v_j)$.
Before that,
we first define a normalized cosine distance %(range from 0 to 1) 
as follows, which is used to measure the similarity of the embedding vectors: %  between $v_i$ and $v_j$

%	\setlength\abovedisplayskip{1.5pt}
%	\setlength\belowdisplayskip{1.5pt}
% 	\begin{small}
		\begin{equation}
		\label{Eq:cosine_similarity}
		{sim}(i,j) =\frac{1}{2}( 1 + \frac{{\bf v}_i\cdot{\bf v}_j}{\parallel {\bf v}_i \parallel \parallel{\bf v}_j \parallel}),
		\end{equation}
% 	\end{small}
where ${\bf v}_i$ is the embedding vector of node $v_i$.
Then,
instead of a naive treatment which rigidly sets $D_{ij}$ as the target similarity and minimizes the error loss between $D_{ij}$ and ${sim}(i,j)$,
we propose a new loss % measure based on the cross-entropy loss, which is used
to measure the similarity cost of each pair of nodes $(v_i,v_j)$ in the embedding space:
% {
%	\vspace{-0.00in}
%	\setlength\abovedisplayskip{1.5pt}
%	\setlength\belowdisplayskip{1.5pt}
% 	\begin{small}
		\begin{equation}
		\label{Eq:similarity_cost}
		l(i,j) = \left\{
		\begin{aligned}
		& -D_{\!ij}\log {sim}(i,j),\ \ \  & D_{ij}>0 \\
		& -\log (1\!- \! {sim}(i,j)),\ \ \ & D_{ij}=0
		\end{aligned}
		\right..
		\end{equation}
% 	\end{small}
% }
%
%
%To achieve this goal,
%we carefully design a new loss % measure based on the cross-entropy loss, which is used
%to measure the similarity cost of each pair of nodes $(v_i,v_j)$ in the embedding space:
%{
%	\setlength\abovedisplayskip{1.5pt}
%	\setlength\belowdisplayskip{1.5pt}
%	\begin{small}
%	\begin{equation}
%	\label{Eq:similarity_cost}
%	l(i,j) = \left\{
%	\begin{aligned}
%	& -D_{\!ij}\log {sim}(i,j),\ \ \  & D_{\!ij}>0 \\
%	& -\log (1\!- \! {sim}(i,j)),\ \ \ & D_{\!ij}=0
%	\end{aligned}
%	\right.,
%	\end{equation}
%	\end{small}
%}
%where ${sim}(\!i,\!j)$ is the normalized cosine similarity score 
%(range from 0 to 1) of the embedding vectors of $v_i$ and $v_j$:
%{
%	\setlength\abovedisplayskip{1.5pt}
%	\setlength\belowdisplayskip{1.5pt}
%	\begin{small}
%		\begin{equation}
%		\label{Eq:cosine_similarity}
%		{sim}(i,j) =\frac{1}{2}( 1 + \frac{{\bf v}_i\cdot{\bf v}_j}{\parallel {\bf v}_i \parallel \parallel{\bf v}_j \parallel}).
%		\end{equation}
%	\end{small}
%}
%where ${\bf v}_i$ is the embedding vector of node $v_i$.
%
There are two subtleties in our careful design of Eq. (\ref{Eq:similarity_cost}). 
First, 
%As Dij has already finished the (process of) normalization, it is still difficult to conclude thta Dij and sim(i,j) have the same dimension, that is the reason why we abandon the method decisely.
%it may be  
%though the proximity is under normalization, it is still difficult to affirm a proximity has the same weight (or measurement) as a similarity score.
%That is, 
%it may be more sensible/reasonable and more flexible to only conclude that larger proximity implies more similarity instead of rigidly think the proximity is exactly the target similarity, and that is why we abandon the aforementioned naive treatment.
%Instead, we impose a penalty to ...
%measurement
%strict
%rigid
% \textcolor{blue}{
though under normalization, one proximity may have a different weight from one similarity score, 
% }
which means rather than treating proximity as the exact target similarity, it can only be concluded that the larger proximity is, the more similarity is.
%leading us to abandon the aforementioned naive treatment.
And that is why we abandon the aforementioned naive treatment.
%Therefore, 
In our design,
%instead of setting the proximity $D_{ij}$ as a stiff target and minimizing the difference between $D_{ij}$ and ${sim}(i,j)$,
we impose a penalty to push $v_i$ and  $v_j$ embedded similarly and set $D_{ij}$ as the penalty coefficient to guarantee that larger proximity will incur more penalty and thus generate a stronger push to be similar.
Second, there is an exception that zero proximity expresses the dissimilarity, which is essentially different from positive proximity. Thus we separate out the zero proximity and set an opposite penalty for it, i.e., impose a penalty %if the nodes with zero-proximity be embedded closely.
to push $v_i$ and  $v_j$ embedded dissimilarly.

%Simultaneously, the design of Eq. (\ref{Eq:similarity_cost}) follows three principles.
%First, the proximity $D_{ij}\!\!>\!\!0$ indicates that $v_i$ and $v_j$ have a similarity, thus we impose a penalty if their similarity score ${sim}(\!i,\!j)$ is low (i.e., they are mapped far away in the embedding space);
%Second, larger proximity  implies more similarity, thus we set the proximity as the penalty coefficient to guarantee that it will incur more penalty if the nodes with larger proximity be embedded far away.
%Third, 
%essentically different from the positive-proximity,
%$D_{ij}\!\!=\!\!0$  expresses the dissimilarity between $v_i$ and $v_j$, thus we separate out the zero-proximity and oppositely impose a penalty if the similarity score ${sim}(\!i,\!j)$ is high.

%
%first, loosen the target proximity as a penalty coefficient; % to weight the similarity cost;
%second, separate zero-proximity from positive-proximity.
%For the first principle,

Then,
by using the loss of Eq. (\ref{Eq:similarity_cost}) for all pairs of nodes, the objective function to preserve the local similarity is defined as follows:
% {
%	\setlength\abovedisplayskip{1.5pt}
%	\setlength\belowdisplayskip{1.5pt}
% 	\begin{small}
		\begin{equation}
		\label{Eq:local_loss}
		\mathcal{L}_{l}=\!\!\!\!\!\!\!\!
		\sum_{i,j \in \{\!D_{ij}\!>\!0\}\!} \!\!\!\!\! \left( -\!D_{ij}\log {sim}(i,j) \right) +\!\!\!\!\!\!\!\!
		\sum_{i,j \in \{\!D_{ij}\!=\!0\}\!} \!\!\!\!\! \left( -\!\log (1\!-\!{sim}(i,j)) \right).
		\end{equation}
% 	\end{small}
% }
Minimizing % the objective function of 
Eq. (\ref{Eq:local_loss}) pushes the similarity of the embedding vectors of $v_i$ and $v_j$ towards $1$ if proximity $D_{ij}$ is large (the larger the proximity, the stronger the push) and pushes it towards $0$ if $D_{ij}=0$.
As a result, we preserve the local pairwise similarity between nodes.

%\vspace{-0.04in}
\subsubsection{Global Listwise Equivalence}
In addition to the local similarity, 
we can extract more
information by considering the relative equivalence. % instead of absolute strength.
For instance,
for three nodes $v_i,v_j,v_k$ with $D_{ij}\!=\!D_{ik}\!=\!D_{jk}\!=\!0.1$,
the local similarity between $v_i$ and $v_j$ may be very weak in view of the strength of $D_{ij}$.
However, 
in a relative view,
we can conclude that $v_i$ and $v_j$ is equivalent in $v_k$'s viewpoint, 
because they distribute the equal proximity with $v_k$. %(i.e., $D_{ik}\!=\!D_{jk}$).
Further, if for the listwise proximity distribution $D_i\!\!=\!\!(D_{i,1},D_{i,2},\cdots\!,D_{i,|V|})$ and $D_j\!\!=\!\!(D_{j,1},D_{j,2},\cdots\!,D_{j,|V|})$,
$v_i$ and $v_j$ always have the above equivalence (i.e., $D_{ik}\!\! \equiv \!\! D_{jk}$ for $\forall k\!\!=\!\!1,\cdots\!,|V|$),
we can conclude that $v_i$ and $v_j$ play an equivalent global structure role in the whole network.
We refer to the above information as the global listwise equivalence.
Note that some works \shortcite{LINE2015} extract similar information from first-order proximity.
We differentiate them in the aspects where we provide a more definite meaning and generalize it as a basic attribute of arbitrary higher-order proximity.

\begin{figure}[t]
% 	\small
% 	%	\vspace{-0.1in}
% 	\setlength{\abovecaptionskip}{-0.1cm}
% 	\setlength{\belowcaptionskip}{-0.2cm}
	\centering
	\includegraphics[width=0.8\textwidth]{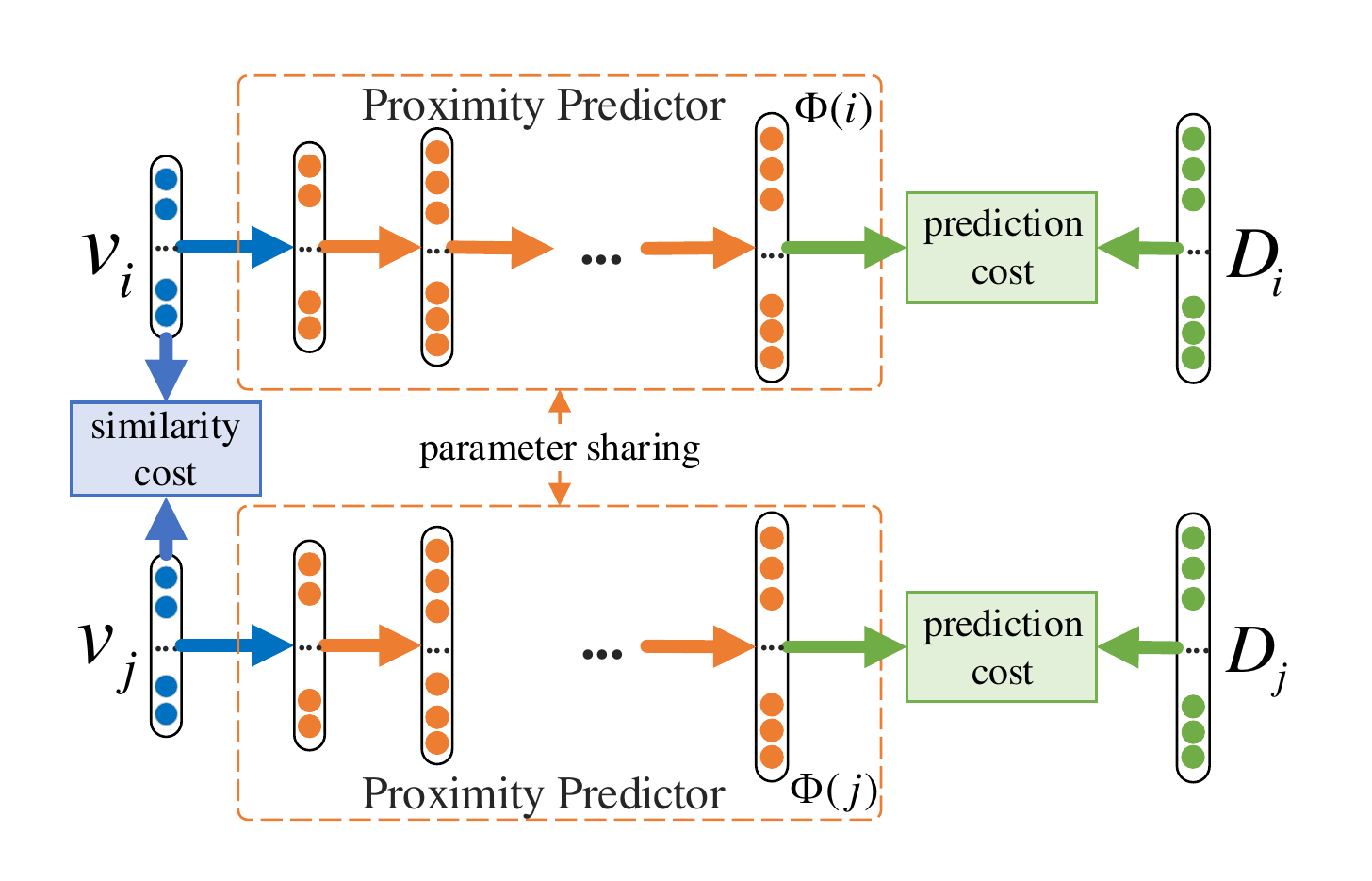}
	\caption{An illustration  of the framework of {\it RWNE} without random walk simulation.}
	\label{Fig:framework}
% 	\vspace{-0.1in}
\end{figure}

To preserve the global listwise equivalence,
%we propose a simple way to directly model the proximity distribution.
%In detail,
%as shown in Figure \ref{Fig:framework},
we propose a self-supervised component: the proximity predictor $\Phi(\cdot)$,
which is a deep architecture composed of multiple nonlinear functions to predict the proximity distribution of an input node.
As shown in Figure \ref{Fig:framework},
for each node $v_i$, the predictor $\Phi(\cdot)$ uses the embedding vector ${\bf v}_i$ as the input and output a $|V|$-dimensional distribution $\Phi(i)\!\in \!\mathbb{R}^{\!1\!\times \!|V|}$, which is the predicted approximation of proximity distribution $D_i$.
%where the $j$-th entry $\Phi(i)_j$ represents the predicted proximity  of $v_j$ in $v_i$'s proximity distribution.
Then, 
supervised by $D_i$, the predictor $\Phi(\cdot)$ is trained to make the output $\Phi(i)$ be close to $D_i$,
which means that if we pick two nodes $v_i$ and $v_j$ with similar proximity distributions (i.e. $D_i\!=\!D_j$),
the predictor $\Phi(\cdot)$ will be trained to learn similar outputs (i.e. $\Phi(i)\!=\!\Phi(j)$) and thus to push the input embedding vector ${\bf v}_i$ and ${\bf v}_j$ to be similar.
Therefore, 
by modeling the proximity in this way, we can learn similar embeddings for nodes with similar proximity distributions.
That is, we capture the global listwise equivalence between nodes.

To train the predictor $\Phi(\cdot)$, 
we follow the inspiration of the design in Eq. (\ref{Eq:similarity_cost}) which sets the target proximity as a penalty coefficient and separates zero proximity from positive proximity,
and design the loss to measure the  prediction cost for each node $v_i$ as follows:
% {
%	\setlength\abovedisplayskip{1.5pt}
%	\setlength\belowdisplayskip{1.5pt}
% 	\begin{small}
		\begin{equation}
		\label{Eq:prediction_cost}
		\begin{aligned}
		\!\!\!g(i) 
		=& -D_i \left[D_{i}>0\right]   \log \Phi(i)   
		-{\bf 1} \cdot \left[D_{i}=0\right] \cdot \log (1\!-\!\Phi(i))  \\
		\!\!\!
		=&\!\!\sum_{j \in \{D_{ij}>0\}} \!\!\!\! \left( -\!D_{ij}\log \Phi(i)_j \right) +  \sum_{j \in \{D_{ij}=0\}} \!\!\!\! \left( -\!\log (1\!-\!\Phi(i)_j) \right).
		\end{aligned}
%		\mathcal{L}_{g} \!\!=\!\! \sum_{j}^{|V|}\!\! \left( \sum_{i \in \{\!D_{\!ij}\!>\!0\}\!} \!\!\!\! \left( -\!D_{\!ij}\log \Phi(j)_i \right)  \!+\!\!\!
%		\sum_{i \in \{\!D_{\!ij}\!=\!0\}\!} \!\!\!\! \left( -\!\log (1\!-\!\Phi(j)_i) \right)\!\!
%		\right)\!. \!\!
		\end{equation}
% 	\end{small}
% }
Note that in principle, 
$\Phi(\cdot)$ can be an arbitrarily deep neural network,
but as we focus on the effort of random walk in this paper, we only use a simple single-layer architecture and leave a superior trying as future work.
In this setting,  the proximity predictor $\Phi(\cdot)$ is defined as:
% {
%	\setlength\abovedisplayskip{1.5pt}
%	\setlength\belowdisplayskip{1.5pt}
% 	\begin{small}
		\begin{equation}
		\label{Eq:fully_nn}
		\Phi(i)=\sigma({\bf W}{\bf v}_i),
		\end{equation}
% 	\end{small}
% }
where ${\bf W}\in\mathbb{R}^{|V|\times d}$,  
%$|\!V\!|$ is the number of nodes, 
$V$ is the nodes set,
$d$ is the embedding dimensionality,
%${\bf v}_j$ is the embedding vector of node $v_j$ and $d$ is the embedding dimensionality.
and $\sigma(\cdot)$ is the logistic function.
%%which is widely adopted in state-of-the-art embedding models~\shortcite{DeepWalk2014,LINE2015,Node2Vec2016}.

Finally,
by using the loss of Eq. (\ref{Eq:prediction_cost}) for all nodes, the objective function to preserve the global equivalence is defined as follows:
% {
%		\vspace{-0.15in}
%	\setlength\abovedisplayskip{1.5pt}
%	\setlength\belowdisplayskip{1.5pt}
% 	\begin{small}
		\begin{equation}
		\label{Eq:global_loss}
		\mathcal{L}_{g} = \sum_{i}^{|V|} \left(  
		\sum_{j \in \{D_{ij}>0\}} \!\!\!\! \left( -\!D_{ij}\log \Phi(i)_j \right) + \!\!\!\! \sum_{j \in \{D_{ij}=0\}} \!\!\!\! \left( -\!\log (1\!-\!\Phi(i)_j) \right)
		\right). 
		\end{equation}
% 	\end{small}
% 	\vspace{-0.1in}
% }

%\vspace{-0.1in}
%\subsubsection{Combining Local Similarity and Global Equivalence}
\subsubsection{The Joint Objective}
To simultaneously preserve both the local pairwise similarity and the global listwise equivalence provided by the higher-order proximity $D$,
we jointly minimize the following objective function, which combines Eq. (\ref{Eq:local_loss}) and Eq. (\ref{Eq:global_loss}):
% {
%		\vspace{-0.1in}
%	\setlength\abovedisplayskip{1.5pt}
%	\setlength\belowdisplayskip{1.5pt}
% 	\begin{small}
		\begin{equation}
		\label{Eq:model_loss_matrix}
		\mathcal{L} = \gamma \mathcal{L}_{l} + \lambda \mathcal{L}_{g} = 
		\sum_{i}^{|V|}  \left(\sum_{j \in \{D_{ij}>0\}} \!\! D_{ij}\ell(i,j)  + \!\!\!\!
		\sum_{j \in \{D_{ij}=0\}} \!\! \zeta(i,j)\!\! \right),
		\end{equation}
% 	\end{small}
% %	\vspace{-0.05in}
% }
where $\ell(i,j)\!=\!-(\gamma \log {sim}(i,j) \!\! + \!\! \lambda \log \Phi(i)_j)$ is the loss for positive proximity,
%is the positive loss that pushs nodes closely embedded,
$\zeta(i,j)\!\! = \!\! -(\gamma \log (1\!-\!{sim}(i,j)) \! + \! \lambda \log (1\!-\!\Phi(i)_j)$  is the loss for zero proximity;
$\gamma$ and $\lambda$ are hyper-parameters to reflect user’s emphasis.

%we first describe the {\it RWNE} model to directly preserve a general form of higher-order proximity $D$ with arbitrary weights, %without any optimization,
%in which we carefully design a sound objective to explicitly capture both the local pairwise similarity and global listwise equivalence between nodes (the framework is shown in Figure \ref{Fig:framework}).

%\vspace{-0.05in}
\subsection{Optimization with Random Walk Simulation}
In practice, an accurate computation of Eq. (\ref{Eq:higher_order_proximities}) to get the higher-order proximity matrix $D$ with $k\geq 2$ is both time and space consuming. % and the complexity increases as $k$ grows.
Thus it is undesirable to directly solve the objective of Eq. (\ref{Eq:model_loss_matrix}) due to the efficiency issue, especially for large-scale networks. % in real world.

To address the above issue,
we highlight that $D$ is essentially an integrated probability matrix 
combining the transition probabilities of a random walk in $1$-st, $2$-nd, $\cdots$, $k$-th step with corresponding weights $\beta_1,\beta_2,\cdots,\beta_k$,
where each entry $D_{ij} = \beta_1 A_{ij} + \beta_2 A_{ij}^2 + \cdots + \beta_k A_{ij}^k$
represents the weighted average hitting probability from $v_i$ to $v_j$ within a $k$-steps random walk.
Then,
without calculating Eq. (\ref{Eq:higher_order_proximities}),
we can invent a $k$-steps ``drop-out'' random walk to simulate the probability matrix $D$: 
in $l$-th step (for $\forall l \!=\!1,2,\cdots\!,k$), 
the walker first randomly moves from the current node to an adjacent node as a normal random walk, 
and then randomly 
%drops out the current hitting node with the probability $1\!\!-\!\!\beta_l$ as the dropping probability.
drops out the current-step hitting node with the dropping probability $1\!\!-\!\!\beta_l$.
By this means, the expected hitting probability from $v_i$ to $v_j$ in $k$ steps is exactly $D_{ij}$.
That is to say, 
for each node $v_i$, 
the above $k$-steps ``drop-out'' random walker starting from $v_i$ hits/samples a paired node $v_j$ with the probability $D_{ij}$.
%More specially, 

In our carefully designed objective of Eq. (\ref{Eq:model_loss_matrix}),
the probability $D_{ij}$ is delicately and theoretically arranged as a weight coefficient. %to weight the loss $\ell(i,j)$. % (as shown in Eq. (\ref{Eq:model_loss_matrix})).
Then,
in the probabilistic setting,
we can equivalently treat the probabilistic real weight as a binary weight by node-sampling treatment, with the sampling probability proportional to the original real weight.
That is, 
we can eliminate the $D_{ij}$ in the first term of Eq. (\ref{Eq:model_loss_matrix}) by sampling nodes with the probability $D_{ij}$, which can be simulated by the aforementioned $k$-steps ``drop-out'' random walks starting from $v_i$.
Furthermore,
for the second term $\sum_{j \in \{D_{ij}=0\}} \! \zeta(i,j)$ in Eq. (\ref{Eq:model_loss_matrix}),
as the matrix $D$ is usually very sparse with many zero entries,
we also leverage an uniform-sampling treatment to optimize it.
Overall, 
with the node-sampling treatment, %approximation,  
the objective function of Eq. (\ref{Eq:model_loss_matrix}) is optimized as:
% {
%	\setlength\abovedisplayskip{1.5pt}
%	\setlength\belowdisplayskip{1.5pt}
\vspace{-0.1in}
% 	\begin{small}
\begin{equation}
		\begin{aligned}
		\mathcal{L} & =  
		\sum_{i}^{|V|}  \left(\sum_{t=1}^{T}  \mathbb{E}_{j \thicksim \{D_{i,j}\}_1^{|V|} }  \ell(i,j)   +\!
		\sum_{t=1}^{T} \mathbb{E}_{j' \thicksim {\bf I}/\{D_{i,j'}>0\} }  \zeta(i,j') \right) \\
		 & =
		\sum_{i}^{|V|}  \sum_{t=1}^{T}  \left( \sum_{j\in p_{i}^{1\to k}}  \ell(i,j)   + \!
		\sum_{m=1}^{|p_{i}^{1\to k}|} \mathbb{E}_{j' \thicksim {\bf I}/\{p_{i}^{1\to k}\} }  \zeta(i,j') \right),
		\label{Eq:model_loss_sampling}
		\end{aligned}
\end{equation}
% 	\end{small}
% \vspace{-0.1in}
% }
where 
%$j \thicksim \{\!D_{i,j}\!\}_1^{|\!V\!|}$ means sampling a node $v_j$ with the probability distribution $\{D_{i,1},D_{i,2},\cdots,D_{i,|V|} \}$, and 
$T$ is the sampling-frequency/walk-times
which can be set as the iteration epochs when solving the objective by an iterative algorithm (e.g. SGD);
$p_{i}^{1\to k}$ is the hitting nodes set in a $k$-steps ``drop-out'' random walk starting from $v_i$;
$\mathbb{E}_{j' \thicksim {\bf I}/\{p_{i}^{1\to k}\} }$ means an uniform-sampling with nodes $\{p_{i}^{1\to k}\}$ excluded.
Note that in each walk, as the random walker samples $|p_{i}^{1\to k}|$ nodes,  we also operate uniform-sampling for the equal times to ensure fairness.

The optimized objective of Eq. (\ref{Eq:model_loss_sampling}) replaces the expensive computation of higher-order proximity with random walk simulation,
%to improve the efficiency and scalability.
%is computationally efficient and scalable for large-scale networks.
and thus has superior efficiency and scalability,
%with both linear computation and storage
and can be easily mini-batch-minimized by applying an iterative algorithm. %s like SGD.
So far,
we have theoretically described a general scalable random-walk-based embedding framework.
%which is able to effectively and efficiently preserve the higher-order proximity. % .with user-specific weights.
%In principle,
For arbitrary user-specified weights $\{\beta_1,\beta_2,\cdots,\beta_k,\cdots\}$,
%weighted combination of proximities of different orders with given weights $\{\beta_1,\beta_2,\cdots,\beta_k,\cdots\}$,
the framework is able to effectively and efficiently preserve the weighted combination of different-order proximities by deploying the aforementioned ``drop-out'' random walk.
%In principle,
Reversely, for an arbitrary user-specified random walk,
the framework can also be used  by directly %deploying the user-specified random walk and 
minimizing the Eq. (\ref{Eq:model_loss_sampling}),
However, superior to other random-walk models, the framework explicitly clarifies what and how network proximity is preserved in random-walk structure, that is 
%the user-specific random walk is essentially used to simulate an integrated higher-order proximity matrix which is carefully preserved in the Eq. (\ref{Eq:model_loss_matrix}).
the average hitting/sampling probability in $k$ steps of the user-specified random walk is just the integrated higher-order proximity %to be preserved and is 
which is theoretically preserved by the objective of Eq. (\ref{Eq:model_loss_matrix}).
%For example, it is obvious that by deploying an uniform random walk, the preserved higher-order proximity matrix is the equal-weighted combination of different-order proximities.

\begin{figure}[t]
% 	\small
	%	\vspace{-0.1in
% 	\setlength{\abovecaptionskip}{-0.1cm}
% 	\setlength{\belowcaptionskip}{-0.2cm}
	\centering
	\includegraphics[width=0.8\textwidth]{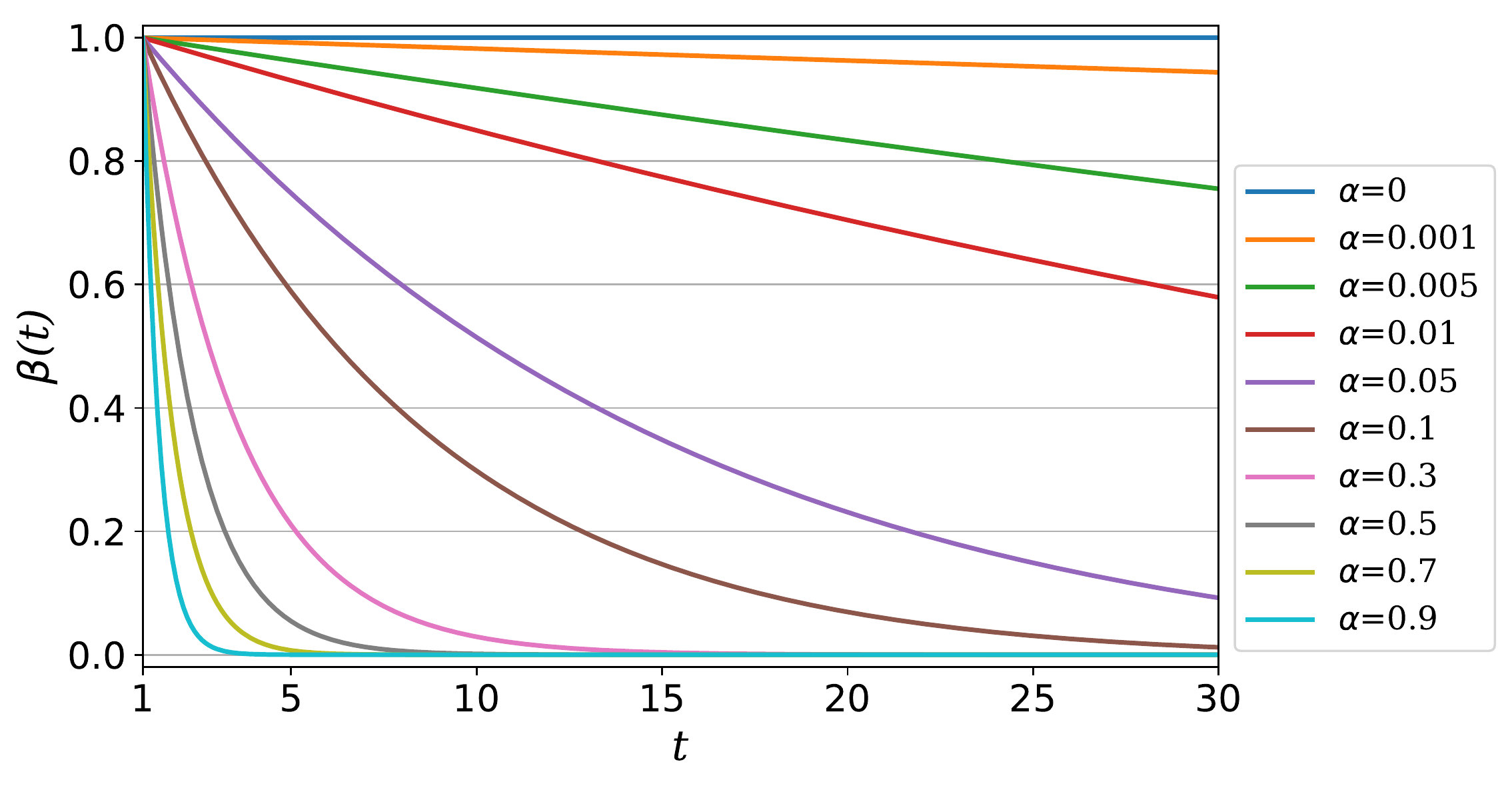}
	\caption{The decline curves of $\beta(t) \!=\! \frac{(1+\alpha (k-t)){(1-\alpha)}^t}{k}$ in different $\alpha$. The $k$ is set as 100 and each curve is normalized by $\beta(1)$.}
	\label{Fig:RWR}
% 		\vspace{-0.1in}
\end{figure}

%\vspace{-0.05in}
\subsubsection{Random Walk with Restart}
%Despite that,
Although we have theoretically described the {\it RWNE} framework,
it is still limited because there is no principled way to determine desirable weights $\{\beta_1,\beta_2,\cdots,\beta_k,\cdots\}$ or random walks.
%To address this issue,
In general,
it is intuitive that the relation/influence over a very long distance can be very weak, i.e., the prestige of different-order proximity is likely to decay with the distance.
Therefore, instead of a normal random walk widely used in the existing models,
we introduce the random walk with restart (referred to as RWR) process:
in each step of a random walk, 
the walker can return back to the root with a personalized teleport probability $\alpha$.
% i.e. a restart probability of returning back to their root.
%in each step, there is a restart probability $\alpha$ that the walker returns back to the root and restarts the walk process, oppositely there is a continue probability $1\!-\!\alpha$ that the walker randomly moves to an adjacent node as a normal random walk process.
The transition probability of RWR is recurrently formalized as:
% {
%\setlength\abovedisplayskip{1.5pt}
%\setlength\belowdisplayskip{1.5pt}
%\vspace{-0.1in}
% \begin{small}
\begin{equation}
\label{Eq:rwr_trainsition}
P^k=\alpha I+ (1-\alpha)P^{k-1}A,\ k=1,2,\cdots,
\end{equation}
% \end{small}
% }
where $P^k$ is the $k$-th step transition probability matrix of RWR, 
and $P^0=I$ is an identity matrix;
$A$ is the single-step  transition probability matrix of normal random walk, i.e., the first-order proximity matrix.
Equivalently, we have:
% {
%\setlength\abovedisplayskip{1.5pt}
%\setlength\belowdisplayskip{1.5pt}
%\vspace{-0.1in}
% \begin{small}
\begin{equation}
P^k={(1\!-\!\alpha)}^k A^k \!+\! \sum_{t=1}^{k}\alpha{(1\!-\!\alpha)}^{k-t} A^{k-t},\ k=1,2,\cdots.
\end{equation}
% \end{small}
% }

Thus, by deploying the RWR process, the integrated higher-order proximity matrix to be preserved (i.e. the average hitting/sampling probability matrix) in $k$ steps is derived as:
% {
%\setlength\abovedisplayskip{1.5pt}
%\setlength\belowdisplayskip{1.5pt}
%\vspace{-0.1in}
% \begin{small}
\begin{equation}\label{Eq:RWR_average_probability}
D = \frac{1}{k}\sum_{t=1}^{k}P^t = \sum_{t=1}^{k} \frac{(1+\alpha (k-t)){(1-\alpha)}^t}{k}  A^t,
\end{equation}
% \end{small}
% }
in which the proximities of different-orders are weighted with a function $\beta(t) \!\!=\!\! \frac{(1+\alpha (k-t)){(1-\alpha)}^t}{k}$.
As Figure \ref{Fig:RWR} shows, 
$\beta(t)$ is approximately an exponentially decreasing function of $t$, since $\alpha\!\! <\!\! 1$.
% \textcolor{blue}{
Compared with other common decay functions, $\beta(t)$ holds the following advantages: (1) The decay rate is first rapid and then slow down as the order $t$ increases, which makes the influence of extreme-high-order proximity smaller. Considering the natural property that the higher-order proximity has less influence in most cases, it is thus reasonable to use $\beta(t)$ instead of the linear or Gaussian decay functions. (2) The decay rate is naturally applied to the RWR process and can be freely adjusted by the personalized teleport probability $\alpha$, while other exponential decay functions are hard to adapt to the random walk process with sufficient theoretical basis.
% }

Finally,
by deploying the RWR process,
{\it RWNE} is able to  effectively and efficiently preserve personalized higher-order proximity, as shown in Eq. (\ref{Eq:RWR_average_probability}) in which the proximities of different orders are naturally weighted with an elegant attenuation %decreasing 
function controlled by a personalized teleport probability.
The final objective of {\it RWNE} by applying the RWR process is as follows:
% {
%	\setlength\abovedisplayskip{1.5pt}
%	\setlength\belowdisplayskip{1.5pt}
%\vspace{-0.1in}
% 	\begin{small}
		\begin{equation}
		\mathcal{L} = \!
		\sum_{i}^{|V|}    \sum_{t=1}^{T}   \left( \sum_{j\in {rwr}_{i}^{1\to k}} \! \!  \ell(i,j) \!  +\!
		\sum_{m=1}^{|{rwr}_{i}^{1\to k}|} \mathbb{E}_{j' \thicksim {\bf I}/\{{rwr}_{i}^{1\to k}\} }  \zeta(i,j')  \right),
		\label{Eq:model_loss_rwr}
		\end{equation}
% 	\end{small}
% }
where ${rwr}_{i}^{1\to k}$ is the hitting nodes set in a $k$-steps random walk with restart process starting from $v_i$;
$\mathbb{E}_{j' \thicksim {\bf I}/\{{rwr}_{i}^{1\to k}\} }$ means an uniform-sampling with nodes ${rwr}_{i}^{1\to k}$ excluded.
%Note that $T$ can be viewed as the iteration epochs of an iterative algorithm (e.g. SGD) which can be applied to minimize the objective. % function.

%Therefore, by applying the node-sampling treatment via RWR, we can not only address the computation problem of high-order proximities, but also naturally obtain a decreasing function to weight different-order proximities.

%\vspace{-0.05in}
\subsection{Complexity Analysis}
\label{sec:comlexity}
In this section, we discuss the time complexity of {\it RWNE}.
We use iterative algorithms (e.g. mini-batch SGD) to minimize the objective function (as shown in Eq. (\ref{Eq:model_loss_rwr})).
In each iteration, we only consider a single root node and deploy a $k$-steps random walk (with restart) starting from it to sample $k$ paired nodes.
As random walk takes only constant time in each step (even in a weighted network, we can use Alias sampling~\shortcite{alias1977} to perform a random walk in $O(1)$ time) and the computation of the loss for each pair of nodes also takes constant time,
we can see that the complexity of each iteration (i.e., the complexity of the inner-body in Eq. (\ref{Eq:model_loss_rwr})) is $O(k)$.
% \textcolor{blue}{
Then,
given the network size $|V|$ (i.e., the number of nodes) and the iteration epochs $T$,
% }
we can extract that the overall time complexity of {\it RWNE} is $O(k \cdot T \cdot |V|)$,
which is linear to the number of nodes in the network.
Therefore, the proposed {\it RWNE} model is computationally efficient and scalable for large-scale networks.

%In practice, we use the asynchronous stochastic gradient descent algorithm (ASGD) to train the model. In each training step, ASGD samples a batch of training node pairs via RWR and simultaneously computes the gradients to update the model parameters. 
%

\subsection{Model Generalization Analysis}
In this section, we extend \emph{RWNE} model as a general framework, in which many popular network embedding approaches are unified.
Under our general framework, it is straightforward to analyze these methods' essences or drawbacks.

\textbf{SpectralClustering}~\shortcite{SpectralClustering2011}:
This method computes the $d$-smallest eigenvectors of normalized Laplacian matrix $L$ as $d$-dimensional network representations, which essentially is a dimensionality reduction based on Laplacian Eigenmaps.
% As SpectralClustering is essentially %substantially
% a dimensionality reduction method based on Laplacian Eigenmaps, it is 
As we also utilize Laplacian Eigenmaps to embed the local pairwise proximity in \emph{RWNE}, it is straightforward that SpectralClustering is a degraded case of \emph{RWNE} by directly decompose the first-order proximity matrix.
Actually, all the network embedding approaches based on the decomposition of adjacent matrix or Laplacian matrix could be substantially regarded as special cases of \emph{RWNE}, which only embed the local pairwise structure information of first-order or higher-order proximity matrix. 
Comparing with direct matrix decomposition, the \emph{RWNE} framework provides a computationally efficient and scalable way to equivalently embed local pairwise proximity.

\textbf{DeepWalk}~\shortcite{DeepWalk2014} and \textbf{node2vec}~\shortcite{Node2Vec2016}:
DeepWalk transforms a network structure into node sequences by random walks, then treats the sequences as ``corpus'' and employs the Skip-gram model for representation learning.
In terms of employing Skip-gram with negative sampling and node-sampling via random walk, DeepWalk is a special case of \emph{RWNE} model which only utilizes the global adjacency proximity between nodes.
% and equivalently treats different order proximities as Eq.\ref{Eq:average_random_walk}.
Similarly, node2vec is also unified into \emph{RWNE} framework by using a second-order random walk to adjust the prestige of different order proximities, which is highly complex and memory-consuming.
However, DeepWalk and node2vec are not direct network embedding methods, as they actually transform the network embedding problem into word embedding problem by preliminarily generating offline ``corpus'' via random walks, which will consume plenty of space, and they do not have explicit objective functions.
The proposed \emph{RWNE} framework do not need to generate a vast ``corpus'', which jointly online walks and trains in each step.
Moreover, \emph{RWNE} framework provides a substantial theoretical foundation and explicit objective for DeepWalk and node2vec.

\textbf{LINE}~\shortcite{LINE2015}:
LINE could be regarded as the first-order case of DeepWalk but directly uses online edge-sampling in each training step instead of generating offline ``corpus''.
Therefore, LINE is also a special case of \emph{RWNE} model which only characterizes the first-order proximity. 

% \textbf{PTE}~\shortcite{PTE2015}: PTE is an extension of LINE (2nd) in heterogeneous text networks, thus is also unified into \emph{RWNE} framework.

% \subsection{Matrix factorization}

%\vspace{-0.05in}
\section{Experiments}\label{sec:experiments}
%\vspace{-0.05in}
In this section, 
% we report experimental results to demonstrate the effectiveness and efficiency of our HPNE model.
we report the experimental results on six real-word datasets to demonstrate the effectiveness and efficiency of our proposed model.
% Code and data will be released in the camera-ready version. % due to the double-blind review policy.

%\vspace{-0.05in}
\subsection{Datasets}\label{sec:datasets}
%\vspace{-0.05in}
%To be convincing and reproducible, here we use the following three publicly available large-scale real-world networks. % (the same as used in DeepWalk).
Here we use the following six publicly available networks with different scales, the statistics of which are shown in Table~\ref{Tab:statistics_of_datasets}.

$\bullet$ \textbf{Cora}~\shortcite{mccallum2000automating} is a widely used scientific publication citation network with 2,708 papers, 5,429 citation links, and  7 categories. The labels represent seven different fields in machine learning research, which are case-based, genetic algorithms, neural networks, probabilistic methods, reinforcement learning, rule learning, and theory.
%It contains 2,708 scientific publications classified into one of seven classes.

%$\bullet$ \textbf{Citeseer}~\shortcite{mccallum2000automating}:
%This is also a citation network with 3,312 nodes, 4,732  links, and  6 categories.

$\bullet$ \textbf{PubMed}~\shortcite{GCN2016} is also a citation network with 19,717 nodes, 44,338  links, and 3 categories.

%$\bullet$ \textbf{PPI}~\shortcite{stark2010biogrid}:
%This is a biological network represents the protein-protein interactions.
%The network has 3,890 nodes, 76,584 edges, and 50 labels.

$\bullet$ \textbf{DBLP}~\shortcite{perozzi2017don} is a well-known academic collaboration network with 27,199 nodes, 133,664 links, and 4 categories.

$\bullet$ \textbf{Blogcatalog}~\shortcite{TangL09_kdd} is a social network of bloggers with 10,312 nodes, 333,983 links, and 39 categories. The labels represent the topic categories provided by the authors.

$\bullet$ \textbf{Flickr}~\shortcite{TangL09_kdd} is a social network of the contacts between users on the Flickr website, with 80,513 nodes, 5,899,882 links, and 195 categories.
The labels represent the interest categories of the users.

$\bullet$ \textbf{Youtube}~\shortcite{TangL09_cikm} is a very large interest network between users on the YouTube website, with 1,138,499 nodes, 2,990,443 links, and 47 labels. The labels represent the interest groups subscribed by users of viewers that enjoy common video genres.

% \begin{table}[t]
% % 	 \vspace{-0.1in}
% 	%\small/
% % 	\footnotesize
% 	\centering
% % 	\renewcommand\arraystretch{0.9}
% % 	\begin{tabular}{p{1.1cm}<{\centering}|p{0.7cm}<{\centering}|p{0.75cm}<{\centering}|p{0.75cm}<{\centering}|p{0.83cm}<{\centering}|p{0.75cm}<{\centering}|p{0.7cm}<{\centering}}
% 	\begin{tabular}{c|r|r|c}
% 		\toprule
% 		Dataset      & nodes & #edges & #categories \\
% 		\midrule
% 		Cora         & 2,780     & 5,429     & 7   \\
% 		\midrule
% 		PubMed       & 19,717    & 44,338    & 3   \\
% 		\midrule
% 		DBLP         & 27,199    & 133,664   & 4   \\
% 		\midrule
% 		Blogcatalog  & 10,312    & 333,983   & 39  \\
% 		\midrule
% 		Flickr       & 80,513    & 5,899,882 & 195 \\
% 		\midrule
% 		Youtube      & 1,138,499 & 2,990,443 & 47 \\
% 		\bottomrule
% 	\end{tabular}
% % 	\vspace{-0.1in}
% 	\caption{Statistics of Datasets.}
% 	\label{Tab:statistics_of_datasets}
% % 	 \vspace{-0.05in}
% \end{table}

\begin{table}[t]
% 	 \vspace{-0.1in}
	%\small/
% 	\footnotesize
	\centering
% 	\renewcommand\arraystretch{0.9}
% 	\begin{tabular}{p{1.1cm}<{\centering}|p{0.7cm}<{\centering}|p{0.75cm}<{\centering}|p{0.75cm}<{\centering}|p{0.83cm}<{\centering}|p{0.75cm}<{\centering}|p{0.7cm}<{\centering}}
	\begin{tabular}{lrrrrrr}
		\toprule
		{\bf Dataset}     & Cora  & PubMed & DBLP    & Blogcat. & Flickr    & Youtube   \\
		\midrule
		{\bf \#nodes}      & 2,780 & 19,717 & 27,199  & 10,312   & 80,513    & 1,138,499 \\
		\midrule
		{\bf \#edges}      & 5,429 & 44,338 & 133,664 & 333,983  & 5,899,882 & 2,990,443 \\
		\midrule
		{\bf \#classes} & 7     & 3      & 4       & 39       & 195       & 47        \\
		\bottomrule
	\end{tabular}
% 	\vspace{-0.1in}
	\caption{Statistics of Datasets.}
	\label{Tab:statistics_of_datasets}
% 	 \vspace{-0.05in}
\end{table}

%\vspace{-0.05in}
\subsection{Baseline Methods and Experimental Settings}
%\vspace{-0.05in}
\label{sec:settings}
We compare our {\it RWNE} model with the following
random-walk-based (DeepWalk, LINE, and node2vec), matrix-factorization-based (GraRep and HOPE), and deep-learning-based (SDNE, GCN, and GAT) network embedding methods, with the hyper-parameters setting listed in Table~\ref{Tab:hyperparameter}.

$\bullet$ \textbf{DeepWalk}~\shortcite{DeepWalk2014}
adopts uniform random walk to capture the contextual information and Skip-gram model to learn node embeddings.
%DeepWalk is a recently proposed network embedding model, which learns $d$-dimensional embeddings by capturing the contextual information via uniform random walks.

$\bullet$ \textbf{LINE}~\shortcite{LINE2015} defines loss functions to preserve 1st- and 2nd- order proximity separately. After optimizing the
loss functions, it concatenates these embeddings.
% LINE is a method that preserves 1st-order and 2nd-order proximity separately.
We use the suggested version to learn two $d/2$-dimensional vectors (one for each-order) and then concatenate them.

$\bullet$ \textbf{node2vec}~\shortcite{Node2Vec2016} is generalized from DeepWalk by introducing a biased random walk.
%node2vec is generalized from DeepWalk, which learns $d$-dimensional embeddings by deploying a  biased-random walk to balance the breadth-first % (BFS) 
%and depth-first % (DFS) 
%graph searches. 

$\bullet$ \textbf{GraRep}~\shortcite{GraRep2015} factorizes the higher-order proximity matrix via
SVD decomposition to get low-dimensional node representations.

$\bullet$ \textbf{HOPE}~\shortcite{HOPE2016} also preserves higher-order proximity based on 
%matrix factorization via generalized SVD.
generalized SVD decomposition.

$\bullet$ \textbf{SDNE}~\shortcite{SDNE2016} uses deep auto-encoders to jointly preserve 1st- and 2nd- order proximity.
%SDNE uses deep auto-encoders to embed network nodes and learn $d$-dimensional embeddings by jointly optimizing two proximities similar to those defined in LINE.

$\bullet$ \textbf{GCN}~\shortcite{GCN2016} is a semi-supervised feature learning model which defines a convolution operator to directly operate graph-structured data.

$\bullet$ \textbf{GAT}~\shortcite{GAN2017} is a novel neural network architecture that operates graph-structured data by leveraging masked self-attentional layers.

% Note that 
%We exclude those matrix-based approaches including classical methods (e.g., LLE~\shortcite{roweis2000nonlinear}) and recent models (e.g., GraRep~\shortcite{GraRep2015}), which are unable to efficiently deal with large-scale networks and have already been shown to be inferior to our baselines.

\textbf{Experimental Setting}. We evaluate the quality of the embedding vectors learned by different methods on three classical network mining
tasks: multi-label node classification, node clustering, and link reconstruction.
To ensure the significance of the results, we repeat each experiment 10 times and %use the mean value as the reporting result. 
report their mean value and standard deviation value.
\begin{table}[t]
% 	\small
	%		\footnotesize
% 	\vspace{-0.15in}
	\centering
	\begin{tabular}{llll}
		\toprule
		Model & Notation & Meaning & Value \\
		\midrule
		\multirow{4}{*}{Common} & 
		$lr$  & learning rate       & 0.025 \\ & 
		$d$   & embedding dimension & 128 \\ & 
		$m$   & negative samples    & 5 \\ & 
		$T$   & iteration epochs    & 20,000 \\
		\midrule
		\multirow{4}{*}{\shortstack{DeepWalk \\  \\ node2vec}} & 
		$wt$   & walk times & 10 \\ & 
		$wl$   & walk length & 80 \\ & 
		$w$    & window size & 10 \\ & 
		$p, q$ & bias parameters of node2vec & 0.25 \\
		\midrule
		\multirow{3}{*}{SDNE} & 
		$\alpha$ & loss weight & 1e3 \\ &
		$\beta$  & non-zero elements weight & 10   \\ &
		$\nu$    & regularizer term weight & 1e-4 \\
		\midrule
		\multirow{4}{*}{\textbf{RWNE}} & 
		$\gamma$  & loss weight & 10 \\ & 
		$\lambda$ & loss weight & 1  \\ & 
		$\alpha$  & personalized teleport probability & 0.3 \\ & 
		$k$       & walk steps & 10 \\
		\bottomrule
	\end{tabular}
	\caption{Hyperparameters setting.}\label{Tab:hyperparameter}
% 	\vspace{-0.15in}
\end{table}

As shown in Table~\ref{Tab:hyperparameter}, 
for the common hyper-parameters, 
we set learning-rate $lr\!\!=\!\!0.025$, embedding-dimension $d\!\!=\!\!128$, negative-samples $m\!\!=\!\!5$, and iteration-epochs $T\!\!=\!\!20000$ for all methods.
%for a trade-off between the computational time and accuracy.
Specially,
%for random walk based methods (DeepWalk, node2vec),
for DeepWalk and node2vec,
we set walk-times $wt\!\!=\!\!10$, walk-length $wl\!\!=\!\!80$, window-size $k\!\!=\!\!10$, and set the bias parameters of node2vec as $p\!\!=\!\!q\!\!=\!\!0.25$, as recommended in their papers.
For SDNE, we tune its parameters of $\alpha,\beta,\nu$ by using a grid-search strategy, and get $\alpha\!\!=\!\!10^{3},\beta\!\!=\!\!10,\nu\!\!=\!\!10^{-4}$.
For GCN and GAT,
we consistently use the structural features (i.e. adjacent matrix) as the input features.
For other parameters and other baselines, we use the default settings as shown in their original papers.
%Specially,
For our {\it RWNE} model,
we set 
the loss-weight $\gamma \!=\!10$ and $ \lambda\!\!=\!\!1$,
the personalized-teleport-probability $\alpha\!\!=\!\!0.3$, 
the walk-steps (walk-length) $k\!\!=\!\!10$.
Note that the walk-steps $k$ in our model actually delimits an upper bound of the order to be preserved which shares a similar meaning with the window-size in DeepWalk and node2vec.

%For all methods, we set the embedding dimension $d\!\!=\!\!128$ as used in DeepWalk, the number of negative samples $m\!=\!5$, the learning rate $lr\!\!=\!\!0.25$ and exponentially decreasing during the training.
%For our HPNE, we set the restart probability $\alpha\!\!=\!\!0.5$, the upper bound of the order $k\!\!=\!\!10$, the hyper-parameter $\lambda\!\!=\!\!1$ to equivalently combine the local and global losses.
%For SDNE, we tune its hyper-parameters of $\alpha,\beta,\nu$ by using a grid-search strategy, and get $\alpha\!\!=\!\!10^{3},\beta\!\!=\!\!10,\nu\!\!=\!\!10^{-4}$.
%For other baselines, we set the total number of samples as $10$ billion for LINE, set walk times $t\!\!=\!\!10$, walk length $l\!\!=\!\!80$, window size $k\!\!=\!\!10$ for both DeepWalk and node2vec, and set the in-out and return parameters $p\!\!=\!\!0.25,q\!\!=\!\!0.25$ for node2vec, as recommended in their papers.
% For other default parameters:
% the embedding dimensionality $d\!\!=\!\!128$ as used in DeepWalk and node2vec, the learning rate $lr=0.25$ and exponentially decreasing during the training. the mini-batch size of the stochastic gradient descent is set 1 for HPNE and LINE, and 128 for SDNE.

% [!htb]
% Experimental results of multi-label node classification on the BlogCatalog, Flickr, and YouTube datasets.
% The results for different training ratios in Blogcatalog, Flickr, and YouTube datasets.

%\vspace{-0.05in}
 \subsection{Multi-Label Node Classification}
 \label{exp:classification}
% \vspace{-0.05in}
% \subsection{Results and Analysis}
%\subsection{Experimental Results}
For the node classification task, we first learn the node embedding vectors from the full nodes on each dataset, and then use the embedding vectors as input features for a one-vs-rest logistic regression classifier, and use both Macro-F1 score and Micro-F1 score as the metrics for evaluation.
We repeat each classification experiment 10 times and randomly split 50\% of the nodes for training and the other 50\% for testing.
Due to the lack of space, we report the mean Macro/Micro-F1 scores in Table~\ref{Tab:results_Macro_f1} and Table~\ref{Tab:results_Micro_f1}, and report the standard deviations in Appendix~\ref{appendix:more_results}.
Note that we exclude the results of some models on the Youtube dataset because they either fail to terminate in one week (SDNE) or run out of memory (GraRep, HOPE, GCN, GAT).

We can observe that our proposed \emph{RWNE} model consistently and significantly
outperforms all the baselines in both metrics on all datasets. 
For example, 
%with small network size, 
on the Cora dataset,
\emph{RWNE} outperforms them by 0.03--0.22 (relatively 4\%--35\%) in terms of Macro-F1 score and by 0.03--0.17 (relatively 4\%--25\%) in terms of Micro-F1 score;
%with large network size, 
on the BlogCatalog dataset with the larger size, \emph{RWNE} also achieves the gains of 0.03--0.16 (relatively 12\%--109\%) in terms of Macro-F1 score and 0.03--0.19 (relatively 7\%--79\%) in terms of Micro-F1 score.

Specially, by only taking random-walk-based methods (DeepWalk, LINE, and node2vec) into account,
%we can find that RWNE consistently improves the classification performance by around 0.04--0.05 (relatively 18\%--19\%) in terms of Macro-F1 scores and by around 0.03--0.04 (relatively 8\%--9\%) in terms of Micro-F1 scores,
we can find that \emph{RWNE} also consistently improves the classification performance by around 0.02--0.08 in both metric scores on all datasets, which clearly demonstrates that the effectiveness of our proposed novel random-walk-based framework.

%%In more detail,
%More generally,
%we can find that,
%by using the adjacent matrix as the input features, 
%the semi-supervised feature learning models (GCN, GAT) show worse performance than expected, especially on the large datasets (Blogcatalog, Flickr),
%which can be explained by its 
%inability to reuse samples,

More generally,
we can find that,
by deploying sufficient long-distance random walks,
random-walk-based algorithms (DeepWalk, node2vec, and our \emph{RWNE}) can achieve considerable improvements
than matrix-factorization-based and deep-learning-based algorithms,
especially on large-scale datasets.
Besides, it is worth mentioning that by using the adjacent matrix as the input features, 
the semi-supervised feature learning models (GCN and GAT) show worse performance than expected, %especially on the Blogcatalog and Flickr datasets, implying 
which implies that these models may not be suitable to exploit a graph with poor attribute features.

%       DeepWalk  & 0.0021 & 0.0015 & 0.0009 & 0.0008 & 0.0064 & 0.0080 \\
% 		LINE      & 0.0105 & 0.0046 & 0.0035 & 0.0028 & 0.0079 & 0.0085 \\
% 		node2vec  & 0.0030 & 0.0021 & 0.0011 & 0.0009 & 0.0078 & 0.0093 \\
% 		\midrule
% 		GraRep    & 0.0006 & 0.0009 & 0.0012 & 0.0009 & 0.0028 & -- \\
% 		HOPE      & 0.0018 & 0.0015 & 0.0015 & 0.0015 & 0.0031 & -- \\
% 		\midrule
% 		SDNE      & 0.0117 & 0.0114 & 0.0157 & 0.0036 & 0.0088 & -- \\
% 		GCN       & 0.0135 & 0.0120 & 0.0159 & 0.0091 & 0.0056 & -- \\
% 		GAT       & 0.0130 & 0.0152 & 0.0209 & 0.0124 & 0.0092 & -- \\
% 		\midrule
% 		RWNE      & 0.0023 & 0.0016 & 0.0021 & 0.0018 & 0.0072 & 0.0078 \\

\begin{table}[t]
	 \vspace{-0.1in}
	%\small/
% 	\footnotesize
	\centering
% 	\renewcommand\arraystretch{0.9}
% \begin{tabular}{p{1.7cm}|p{1.6cm}<{\centering}|p{1.6cm}<{\centering}|p{1.6cm}<{\centering}|p{1.6cm}<{\centering}|p{1.6cm}<{\centering}|p{1.6cm}<{\centering}}
\begin{tabular}{p{1.7cm}p{1.6cm}<{\centering}p{1.6cm}<{\centering}p{1.6cm}<{\centering}p{1.6cm}<{\centering}p{1.6cm}<{\centering}p{1.6cm}<{\centering}}
% 	\begin{tabular}{l|c|c|c|c|c|c}
		\toprule
		{\bf Dataset}   & {\bf Cora}   & {\bf Pubmed} & {\bf DBLP}   & {\bf Blogcat.} & {\bf Flickr} & {\bf Youtube}  \\
		\midrule
		DeepWalk  & 0.7970 & 0.7764 & 0.5916 & 0.2649  & 0.2659 & 0.3741 \\
		LINE      & 0.7493 & 0.7454 & 0.5702 & 0.2596  & 0.2582 & 0.3463 \\
		node2vec  & 0.7970 & 0.7817 & 0.5920 & 0.2677  & 0.2637 & 0.3766 \\
		\midrule
		GraRep    & 0.7756 & 0.7679 & 0.5683 & 0.2430  & 0.2590 & -- \\
		HOPE      & 0.6156 & 0.6357 & 0.4555 & 0.2008  & 0.1904 & -- \\
		\midrule
		SDNE      & 0.6879 & 0.6700 & 0.5227 & 0.2130  & 0.2332 & -- \\
		GCN       & 0.7390 & 0.7113 & 0.5330 & 0.1429  & 0.1568 & -- \\
		GAT       & 0.7541 & 0.7203 & 0.5513 & 0.1659  & 0.1812 & -- \\
		\midrule
		{\bf RWNE}  & {\bf 0.8318} & {\bf 0.8067} & {\bf 0.6149} & {\bf 0.2992} & {\bf 0.2910} & {\bf 0.3993} \\
		\bottomrule
	\end{tabular}
	\vspace{-0.1in}
	\caption{The Macro-F1 scores for multi-label node classification.}
	\label{Tab:results_Macro_f1}
% 	 \vspace{-0.05in}
\end{table}

%[!htb]
\begin{table}[t]
	 \vspace{-0.05in}
%	\small/
% 	\footnotesize
	\centering
% 	\renewcommand\arraystretch{0.9}
% \begin{tabular}{p{1.7cm}|p{1.6cm}<{\centering}|p{1.6cm}<{\centering}|p{1.6cm}<{\centering}|p{1.6cm}<{\centering}|p{1.6cm}<{\centering}|p{1.6cm}<{\centering}}
\begin{tabular}{p{1.7cm}p{1.6cm}<{\centering}p{1.6cm}<{\centering}p{1.6cm}<{\centering}p{1.6cm}<{\centering}p{1.6cm}<{\centering}p{1.6cm}<{\centering}}
% 	\begin{tabular}{l|c|c|c|c|c|c}
		%\begin{tabular}{c|c|c|c|c|c|c}
		\toprule
		{\bf Dataset}   & {\bf Cora}   & {\bf Pubmed} & {\bf DBLP}   & {\bf Blogcat.} & {\bf Flickr} & {\bf Youtube}\  \\
		\midrule
		DeepWalk  & 0.8085 & 0.8063 & 0.6454 & 0.3922 & 0.3981 & 0.4436 \\
		LINE      & 0.7614 & 0.7551 & 0.6157 & 0.3827 & 0.3739 & 0.4272 \\
		node2vec  & 0.8074 & 0.8060 & 0.6520 & 0.3965 & 0.3962 & 0.4487 \\
		\midrule
		GraRep    & 0.7873 & 0.7808 & 0.6284 & 0.3852 & 0.3801 & -- \\
		HOPE      & 0.6736 & 0.6478 & 0.5683 & 0.3259 & 0.2935 & -- \\
		\midrule
		SDNE      & 0.7311 & 0.7284 & 0.6013 & 0.3520 & 0.3611 & -- \\
		GCN       & 0.7672 & 0.7591 & 0.6262 & 0.2378 & 0.2493 & -- \\
		GAT 	  & 0.7834 & 0.7595 & 0.6387 & 0.2721 & 0.2694 & -- \\
		\midrule
		{\bf RWNE}  & {\bf 0.8430} & {\bf 0.8289} & {\bf 0.6718} & {\bf 0.4247} & {\bf 0.4157} & {\bf 0.4622} \\
		\bottomrule
	\end{tabular}
	\vspace{-0.1in}
	\caption{The Micro-F1 scores for multi-label node classification.}
	\label{Tab:results_Micro_f1}
% 	 \vspace{-0.15in}
\end{table}

%By varying the train-test split ratio, we can observe that the proposed HPNE model consistently outperform all baselines in terms of both metrics and all datasets.
%In Blogcatalog, we improve the classification performance in different train-test ratios by 2\%-5\% over DeepWalk and node2vec, and by 3\%-16\% over LINE and SDNE, respectively.
%In Flickr, the improvement is consistent, on average by 1\%-13\% over baseline models.
%% We observe that SDNE shows much worse performance than expected, which may be explained by its unpractical training requirements.
%% The process of tuning its several hyper-parameters requires efforts and expertise, and when dealing with a large network, the deep neural networks it uses becomes very large and difficult to train.
%% The experiments in YouTube also provide an evidence in a way.
%% In YouTube, the comparison 
%The YouTube network is considerably larger and sparser than the previous ones, and its size prevents SDNE from running on our experimental platform (16GB Tesla K80 GPU). % by using the authors' code.
%However, our method consistently outperform other baselines by 1\%-5\% improvements in both metrics.
%The results demonstrate that the proposed HPNE model can effectively learn high-quality embeddings for large-scale networks.

% preserve high-order proximities and tackle the scalability and sparsity of the network, and thus learn high-quality embeddings. %especially achieving big improvements in large-scale sparse networks.

\clearpage
% be inferior to 
% [!htb]
\begin{table}[t]
% 	 \vspace{-0.1in}
	%\small/
% 	\footnotesize
	\centering
% 	\renewcommand\arraystretch{0.9}
% \begin{tabular}{p{1.7cm}|p{1.6cm}<{\centering}|p{1.6cm}<{\centering}|p{1.6cm}<{\centering}|p{1.6cm}<{\centering}|p{1.6cm}<{\centering}|p{1.6cm}<{\centering}}
\begin{tabular}{p{1.7cm}p{1.6cm}<{\centering}p{1.6cm}<{\centering}p{1.6cm}<{\centering}p{1.6cm}<{\centering}p{1.6cm}<{\centering}p{1.6cm}<{\centering}}
% 	\begin{tabular}{c|c|c|c|c|c|c}
		\toprule
		{\bf Dataset}   & {\bf Cora}   & {\bf Pubmed} & {\bf DBLP}   & {\bf Blogcat.} & {\bf Flickr} & {\bf Youtube}\  \\
		\midrule
		DeepWalk  & 0.4440 & 0.2852 & 0.1811 & 0.1734 & 0.3328 & 0.3093 \\
		LINE  	  & 0.3516 & 0.2302 & 0.1558 & 0.1672 & 0.3197 & 0.2877 \\
		node2vec  & 0.4419 & 0.2871 & 0.1850 & 0.1786 & 0.3374 & 0.3088 \\
		\midrule
		GraRep    & 0.4166 & 0.2587 & 0.1626 & 0.1687 & 0.3202 & -- \\
		HOPE  	  & 0.2986 & 0.2295 & 0.1228 & 0.1387 & 0.2410 & -- \\
		\midrule
		SDNE  	  & 0.3215 & 0.1789 & 0.1385 & 0.1439 & 0.2801 & -- \\
		GCN  	  & 0.3607 & 0.1956 & 0.1316 & 0.1134 & 0.1422 & -- \\
		GAT  	  & 0.3995 & 0.2078 & 0.1451 & 0.1261 & 0.1895 & -- \\
		\midrule
		{\bf RWNE}  & {\bf 0.4683} & {\bf 0.3049} & {\bf 0.1986} & {\bf 0.1953} & {\bf 0.3543} & {\bf 0.3209} \\
		\bottomrule
	\end{tabular}
% 	\vspace{-0.1in}
	\caption{The NMI scores for node clustering.}
	\label{Tab:results_nmi}
% 	 \vspace{-0.10in}
\end{table}

%\vspace{-0.05in}
\subsection{Node Clustering}
For the node clustering task, 
we use the embedding vectors as the input to a $k$-means cluster and evaluate the performance in terms of NMI (Normalized Mutual Information) score.
And also, all the experiments are conducted 10 times, and the mean NMI scores are shown in Table~\ref{Tab:results_nmi}, and the standard deviations are shown in Appendix~\ref{appendix:more_results}.
%Note that we also exclude the results of SDNE, HOPE, GCN, and GAT on the Youtube dataset due to the same reason introduced in Section \ref{exp:classification}.

Overall, the results of node clustering are consistent with the results of node classification, and we can reach a similar conclusion as
analyzed in Section~\ref{exp:classification}.
We can see that the proposed \emph{RWNE} model consistently and clearly outperforms all the comparative baselines on all datasets.
For example, 
%on the Cora dataset, RWNE improves the NMI score by around 0.02--0.24 (relatively 5\%--102\%) over DeepWalk, LINE and node2vec, and by around 0.07--0.32 (relatively 17\%--206\%) over HOPE, SDNE, GCN and GAT;
%on the Blogcatalog dataset, RWNE also achieves the gains by around 0.01--0.07 (relatively 3\%--53\%) over DeepWalk, LINE and node2vec,  and by around 0.10--0.14 (relatively 107\%--301\%) over HOPE, SDNE, GCN and GAT.
on the Cora dataset,
\emph{RWNE} improves the NMI score by 0.02--0.12 (relatively 5\%--33\%) over random-walk-based models, and by 0.05--0.17 (relatively 12\%--57\%) over other baseline models;
on the BlogCatalog dataset,
\emph{RWNE} improves the NMI score by around 0.02--0.03 (relatively 9\%--17\%) over random-walk-based models, and by around 0.03--0.08 (relatively 15\%--72\%) over other baseline models.

%
%\begin{table}
%	% \vspace{-0.1in}
%	%\small/
%	\footnotesize
%	\centering
%	\renewcommand\arraystretch{0.9}
%	\begin{tabular}{p{1.1cm}<{\centering}|p{0.7cm}<{\centering}|p{0.75cm}<{\centering}|p{0.75cm}<{\centering}|p{0.83cm}<{\centering}|p{0.75cm}<{\centering}|p{0.7cm}<{\centering}}
%		%\begin{tabular}{c|c|c|c|c|c|c}
%		\toprule
%		Dataset & Cora & Citeseer & PPI & Blogcat. & Flickr & Youtube  \\
%		\midrule
%		DeepWalk  & 0.4440 & 0.2184 & 0.2228 & 0.1834 & 0.3428 & 0.3193 \\
%		LINE  & 0.2516 & 0.1178 & 0.1932 & 0.1272 & 0.2897 & 0.2277 \\
%		node2vec  & 0.4419 & 0.2171 & 0.2190 & 0.1886 & 0.3424 & 0.3188 \\
%		\midrule
%		HOPE  & 0.1806 & 0.0689 & 0.1471 & 0.0487 & 0.2210 & -- \\
%		SDNE  & 0.3215 & 0.1351 & 0.2093 & 0.0939 & 0.3001 & -- \\
%		GCN  & 0.3607 & 0.1956 & 0.1669 & 0.0634 & 0.1222 & -- \\
%		GAT  & 0.3995 & 0.2286 & 0.1272 & 0.0761 & -- & -- \\
%		\midrule
%		{\bf RWNE}  & {\bf 0.4683} & {\bf 0.2414} & {\bf 0.2430} & {\bf 0.1953} & {\bf 0.3503} & {\bf 0.3209} \\
%		\bottomrule
%	\end{tabular}
%	\vspace{-0.1in}
%	\caption{\small Comparison of running time.}
%	\label{Tab:results_time_consumption}
%	% \vspace{-0.1in}
%\end{table}
%
%\subsection{Time and Space Consumption}
%
%[!htb]

\begin{table}[t]
	%	 \vspace{-0.1in}
	%\small/
% 	\footnotesize
	\centering
% 	\renewcommand\arraystretch{0.9}
% \begin{tabular}{p{1.7cm}|p{1.6cm}<{\centering}|p{1.6cm}<{\centering}|p{1.6cm}<{\centering}|p{1.6cm}<{\centering}|p{1.6cm}<{\centering}|p{1.6cm}<{\centering}}
\begin{tabular}{p{1.7cm}p{1.6cm}<{\centering}p{1.6cm}<{\centering}p{1.6cm}<{\centering}p{1.6cm}<{\centering}p{1.6cm}<{\centering}p{1.6cm}<{\centering}}
% 	\begin{tabular}{l|c|c|c|c|c|c}
        \toprule
        {\bf Dataset}   & {\bf Cora}   & {\bf Pubmed} & {\bf DBLP}   & {\bf Blogcat.} & {\bf Flickr} & {\bf Youtube}\  \\
        \midrule
        DeepWalk  & 0.7849 & 0.6033 & 0.7541 & 0.2415 & 0.2737 & 0.2739 \\
        LINE      & 0.7599 & 0.5746 & 0.7183 & 0.2275 & 0.2570 & 0.2458 \\
        node2vec  & 0.8192 & 0.6177 & 0.7873 & 0.2616 & 0.2883 & 0.2994 \\
        \midrule
        GraRep    & 0.6944 & 0.5853 & 0.7127 & 0.2298 & 0.2522 & -- \\
        HOPE      & 0.6126 & 0.4594 & 0.5909 & 0.1748 & 0.1633 & -- \\
        \midrule
        SDNE      & 0.7387 & 0.5511 & 0.6400 & 0.1565 & 0.1817 & -- \\
        GCN       & 0.6509 & 0.4851 & 0.5581 & 0.1173 & 0.1125 & -- \\
        GAT       & 0.6991 & 0.4891 & 0.5975 & 0.1243 & 0.1332 & -- \\
        \midrule
        {\bf RWNE}      & {\bf 0.8596} & {\bf 0.6435} & {\bf 0.8209} & {\bf 0.2815} & {\bf 0.3129} & {\bf 0.3167} \\
        \bottomrule
	\end{tabular}
% 	\vspace{-0.1in}
	\caption{The MAP scores for link reconstruction.}
	\label{Tab:results_map}
% 	\vspace{-0.1in}
\end{table}

\subsection{Link Reconstruction}
For the link reconstruction task~\shortcite{SDNE2016}, 
we rank pairs of nodes according to their similarities, i.e. the inner product of two embedding vectors, and then reconstruct/predict the links for the highest-ranking pairs of nodes.
We use the MAP (Mean Average Precision) metric~\shortcite{goyal2018graph} to estimate the reconstruction precision.
Also, all the experiments are conducted 10 times, and the mean MAP scores are shown in Table~\ref{Tab:results_map}, and the standard deviations are shown in Appendix~\ref{appendix:more_results}.

We can observe that the results of link reconstruction are also consistent with the results of node classification and node clustering, and we can draw similar conclusions. 
Overall,
in terms of the MAP scores on all the six datasets,
%the proposed RWNE consistently and considerably outperforms DeepWalk and node2vec by around 0.02--0.11 on all the seven datasets, and 
the proposed \emph{RWNE} consistently and substantially achieves around 0.02--0.07 improvements over DeepWalk and node2vec, and around 0.05--0.27 gains over other baselines.

%\vspace{-0.03in}
\subsection{Parameter Sensitivity}
%In this section, we illustrate the parameters sensitivity of the HPNE model in the BlogCatalog dataset.
% we examine how the different choices of parameters affect the performance of HPNE on the BlogCatalog dataset using a 50-50 split between training and testing data.
%For each experiment, we vary one parameter and fix the others, and report the Micro-F1 score by using a train-test split of 50\%-50\%.

\begin{figure}[t]
% 	\small
%	\vspace{-0.1in}
% 	\setlength{\abovecaptionskip}{-0.08cm}
% 	\setlength{\belowcaptionskip}{-0.4cm}
	\centering
	\subfigure[walk steps $k$.]{
		\label{Fig:walk_steps}
		\includegraphics[width=0.45\textwidth]{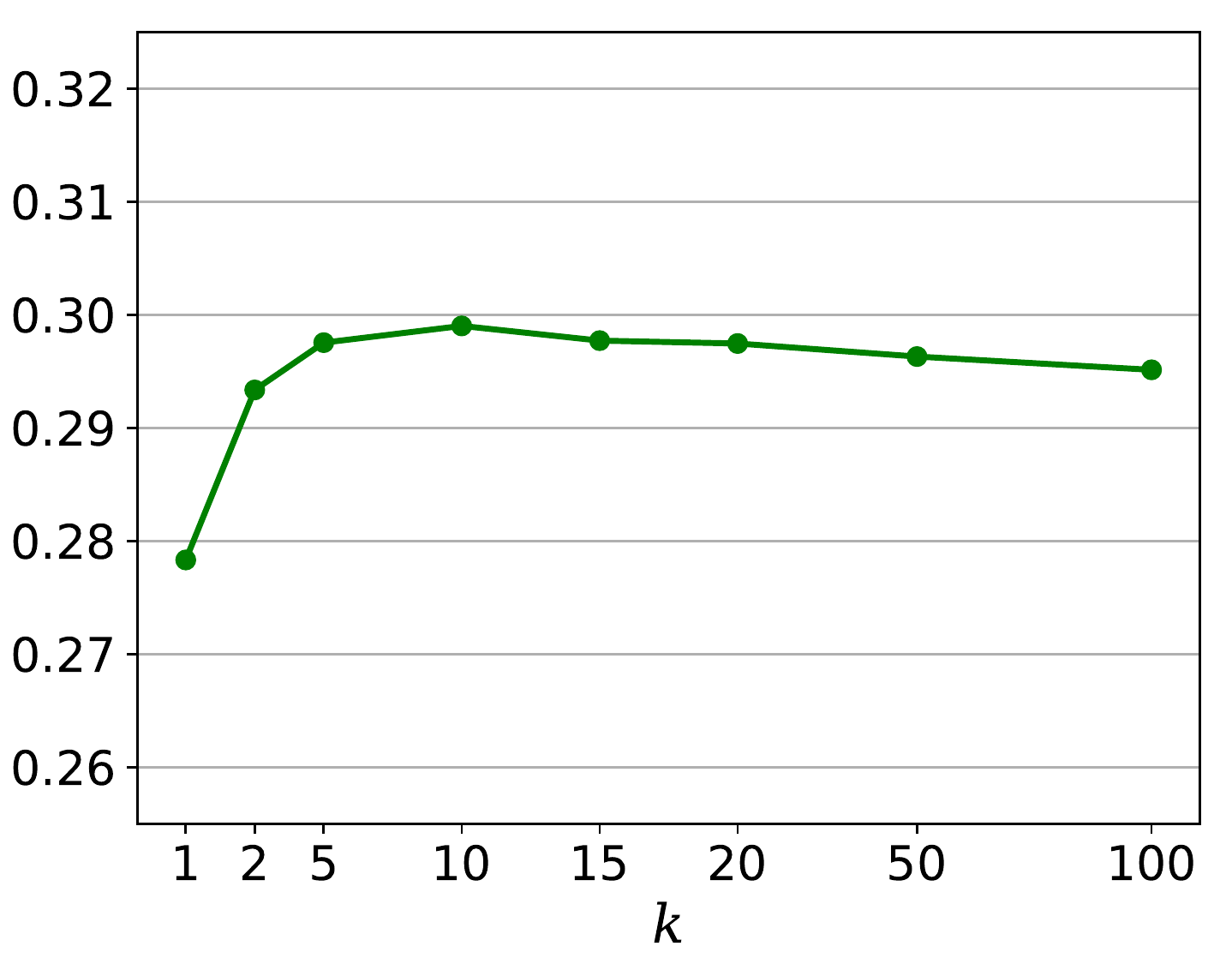}
	}
	%     \hfill
% 	\hspace{-0.1in}
	    \vspace{-0.1in}
	\subfigure[teleport probability $\alpha$.]{
		\label{Fig:teleport_probability}
		\includegraphics[width=0.45\textwidth]{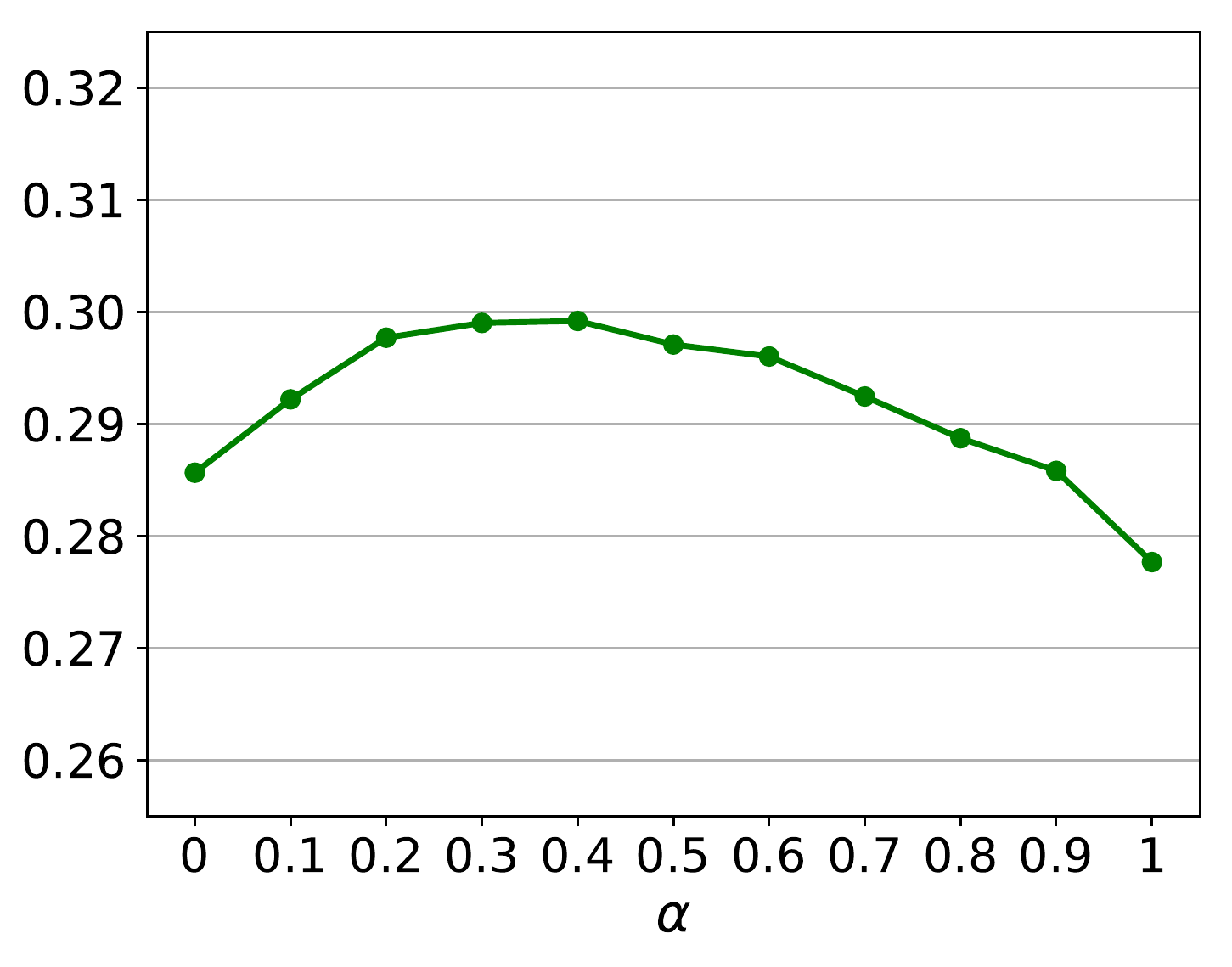}
	}
% 	\hspace{-0.1in}
	    \vspace{0.in}
	\subfigure[loss weight $\gamma$.]
	{\label{Fig:local_loss_factor}
		\includegraphics[width=0.45\textwidth]{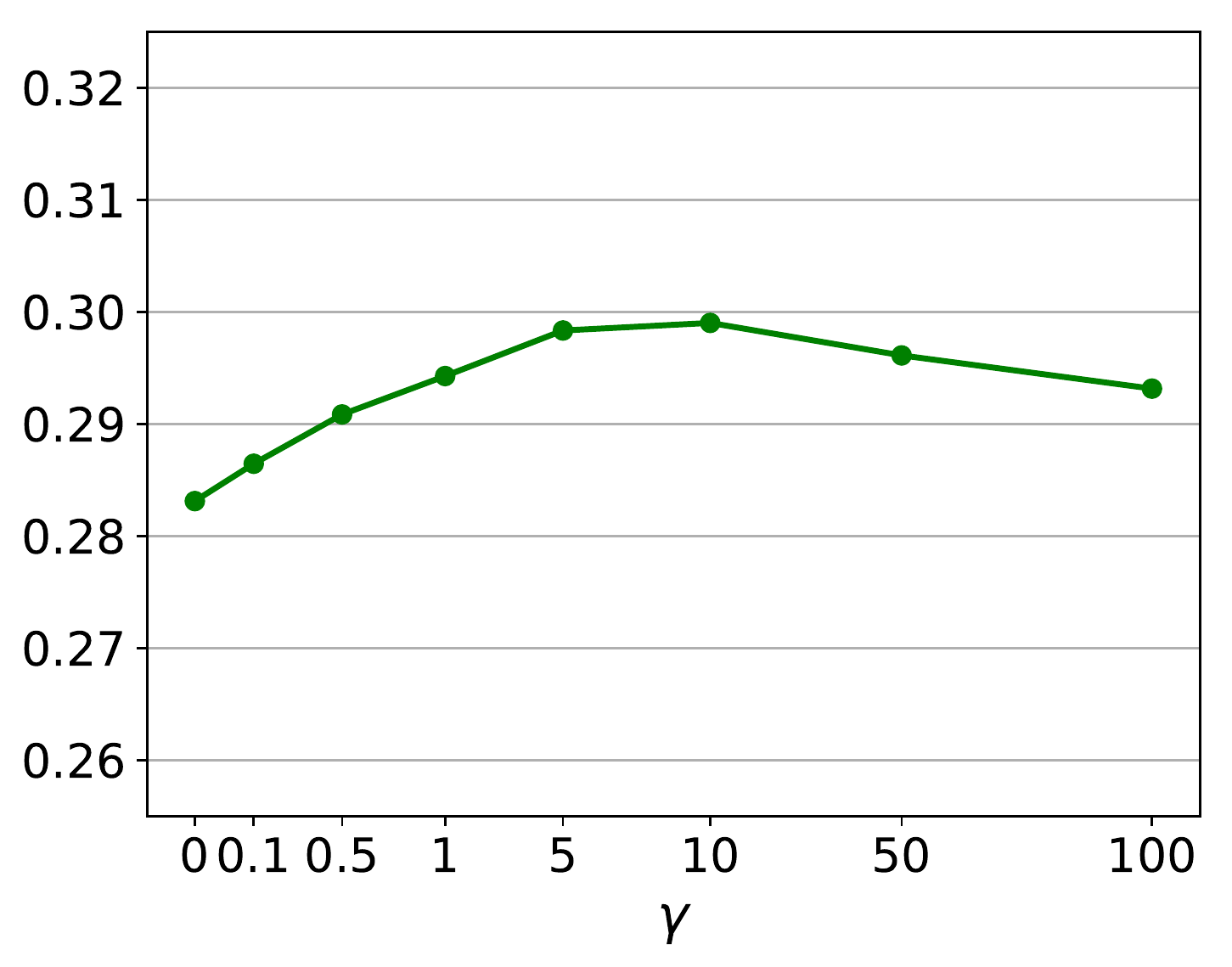}
	}
% 	\hspace{-0.1in}
	    \vspace{0.in}
	\subfigure[loss weight $\lambda$.]
	{\label{Fig:global_loss_factor}
		\includegraphics[width=0.45\textwidth]{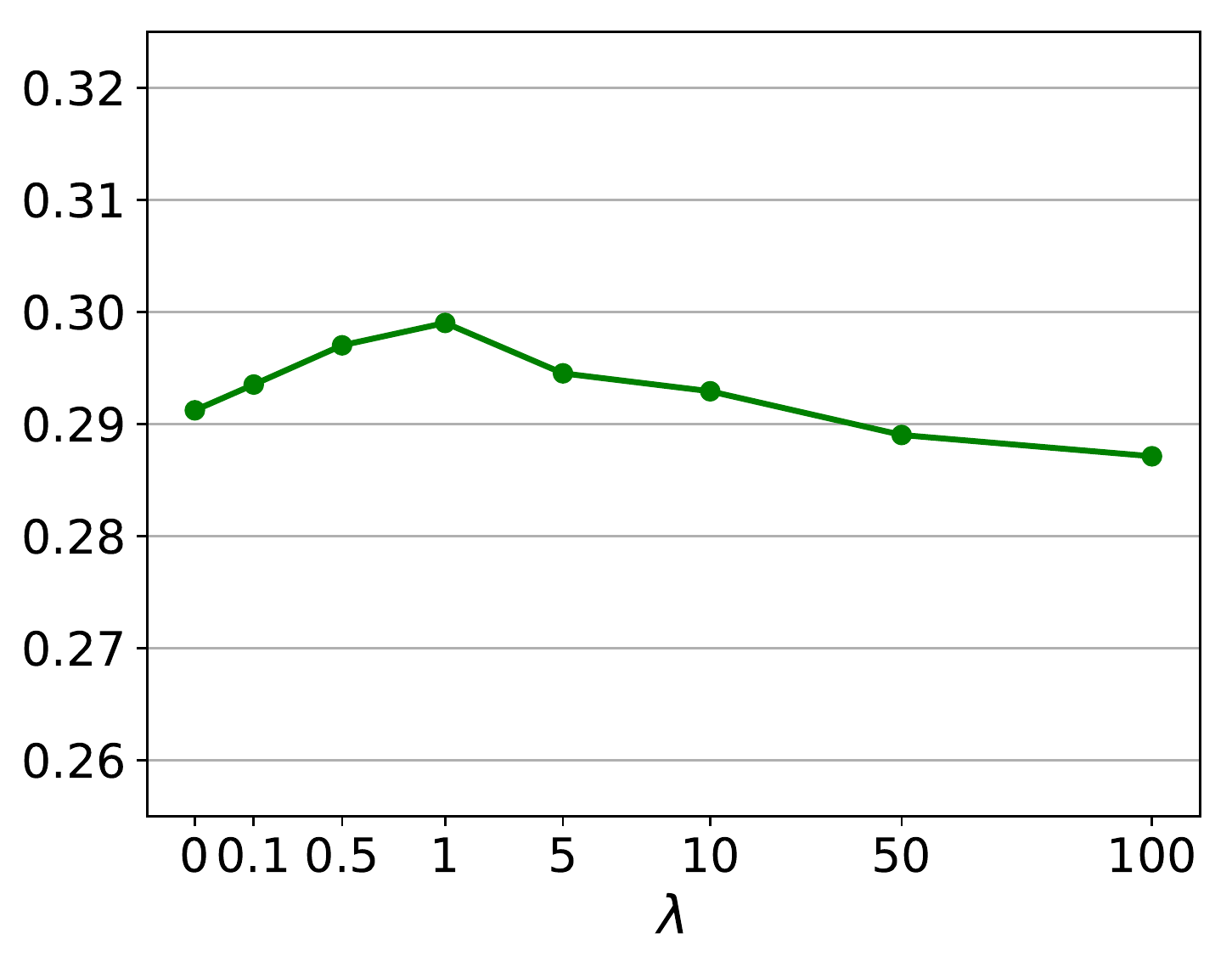}
	}
	\caption{Parameters sensitivity by the Macro-F1 scores of node classification experiments on the Blogcatalog dataset.}
	\label{Fig:Parameter_sensitivity}
	\vspace{0.1in}
\end{figure}

In our model, there exist four important parameters: walk-steps $k$, personalized teleport probability $\alpha$, and loss weight $\gamma$ and $ \lambda$.
In this section, we illustrate these parameters sensitivity by the Macro-F1 scores of node classification experiments on the Blogcatalog dataset.
For each experiment, we vary one parameter and fix the others as the default values (as shown in Section \ref{sec:settings}).
The results are shown in Figure \ref{Fig:Parameter_sensitivity}.

% From Figure \ref{Fig:walk_steps}, 
% %we can see that  when $k\!\!<\!\!10$, the performance improves with $k$,
% %and the improvement seems to be gentle when $k\!\!>\!\!10$, 
% we can see that the performance is soaring when $k\!\!<\!\!10$ while it shows a slight decline trend when $k\!\!>\!\!10$,
% which reveals that it is beneficial to capture the proximity within an appropriate level of scope.
% From Figure \ref{Fig:teleport_probability}, we can observe that the model achieves the best performance when 
% $\alpha$ reaches around 0.3,
% %$\alpha\!\!=\!\!0.3$, 
% which clearly proves the necessity and effectiveness of random walk with restart process.
% In addition,
% Figure \ref{Fig:local_loss_factor} and Figure \ref{Fig:global_loss_factor} show the effects of varying the weight $\gamma$ and $ \lambda$.
% % to weight the losses designed for the pairwise similarity and the listwise equivalence information provided by higher-order proximity (as introduced in Section \ref{local_and_global}).
% We can observe that  it is worthwhile to jointly preserve the local similarity and global equivalence with 
% suitable weights (e.g., $\gamma\!=\!10$ and $ \lambda\!=\!1$).
% %desirable weight to control the proportion (e.g., $\gamma\!\!=\!\!10$ and $ \lambda\!\!=\!\!1$).
% %simultaneously preserve the direct similarity and the relative equivalence, 

\textbf{Walk-Steps $k$}. We first examine the effects of increasing the upper bound $k$ of the order we combine (see the solid line in Figure \ref{Fig:walk_steps}).
We can observe that the model will converge when we consider a large upper bound.
Specially, when $k\!\!<\!\!10$, the performance increases as $k$ increases.
When $k\!\!>\!\!10$, the performance is steady even if we expand $k$ by several times.
Therefore, we can fix $k$ as a reasonably large value in practice, and then we can easily learn a well-performed model by only tuning the restart probability.
In contrast, the amount of similar hyper-parameters % that need be carefully tuned 
are three in DeepWalk and five in node2vec.
The results prove that our model is very practical to achieve better performance with fewer hyper-parameters.
the effectiveness and practicality of the proposed model.

\textbf{Personalized Teleport Probability $\alpha$}. We show the effects of varying the restart probability $\alpha$ of RWR.
As shown in Eq. (\ref{Eq:RWR_average_probability}), 
RWR weights different-order proximities with a decreasing function $\beta(t) \!\!=\!\! \frac{(1+\alpha (k-t)){(1-\alpha)}^t}{k}$.
By varying $\alpha$,  the decreasing rapidity of $\beta(t)$ varies as shown in Figure \ref{Fig:RWR}. 
And correspondingly, the experimental results with different $\alpha$ are shown as the solid line in Figure \ref{Fig:teleport_probability}.
We can observe that the model achieves the optimal performance when $\alpha\!\!=\!\!0.3$. If we vary to a smaller $\alpha$, the performance also falls down. Particularly, when $\alpha\!=\!0$, the model degrades to equivalently treat different-order proximities similar to DeepWalk.
If we choose a bigger $\alpha$, the prestige of higher-order proximities will decline. Extremely, only the first-order proximity can be preserved when $\alpha\!=\!1$, which is clearly insufficient and inferior.

\textbf{Loss Weight $\gamma$ and $ \lambda$}. In addition,
Figure \ref{Fig:local_loss_factor} and Figure \ref{Fig:global_loss_factor} show the effects of varying the $\gamma$ and $ \lambda$ , which are used to weight the losses designed for the pairwise similarity and the listwise equivalence information provided by higher-order proximity (as introduced in Section \ref{local_and_global}).
We can observe that  it is worthwhile to jointly preserve the local similarity and global equivalence with 
suitable weights (e.g., $\gamma\!=\!10$ and $ \lambda\!=\!1$).
%desirable weight to control the proportion (e.g., $\gamma\!\!=\!\!10$ and $ \lambda\!\!=\!\!1$).
%simultaneously preserve the direct similarity and the relative equivalence, 

% We also investigate the effects of varying the parameter $\lambda$ (see in Eq.(\ref{Eq:sour_loss_k_order_all})) to weight the local losses.
% For simplicity, we only report two cases: 
% the solid lines in Figure \ref{Fig:restart} and \ref{Fig:order} show the results 
% with %by considering the
% local losses (i.e., $\lambda\!=\!1$), 
% oppositely the dotted lines in Figure \ref{Fig:restart} and \ref{Fig:order}
% show the results without %by omitting the 
% local losses  (i.e., $\lambda\!=\!0$).
% We can observe that co-optimizing local losses considerably improves the performance when we only consider the first-order proximity. %(less than five order).
% Although the improvement becomes weaker when we consider more high-order proximities, we can also achieve about 1\% gains.
% We believe the main cause is that the global information provided by high-order proximities is quite sufficient to fully model the network structures, which makes the local information less unhelpful.
% The analysis is also consistent with the previous convergence conclusion.
% % Nevertheless, 
% Overall, 
% % it is worthwhile to co-optimize the local losses to ensure high-quality embeddings.
% the results indicate that the local and global proximities we designed are both necessary and helpful to improve the embeddings.

% [!hbt]

\begin{figure}[t]
% 	\small
%		\setlength{\abovecaptionskip}{-0.01cm}
% 		\setlength{\belowcaptionskip}{-0.1cm}
	\centering
	\includegraphics[width=0.5\textwidth]{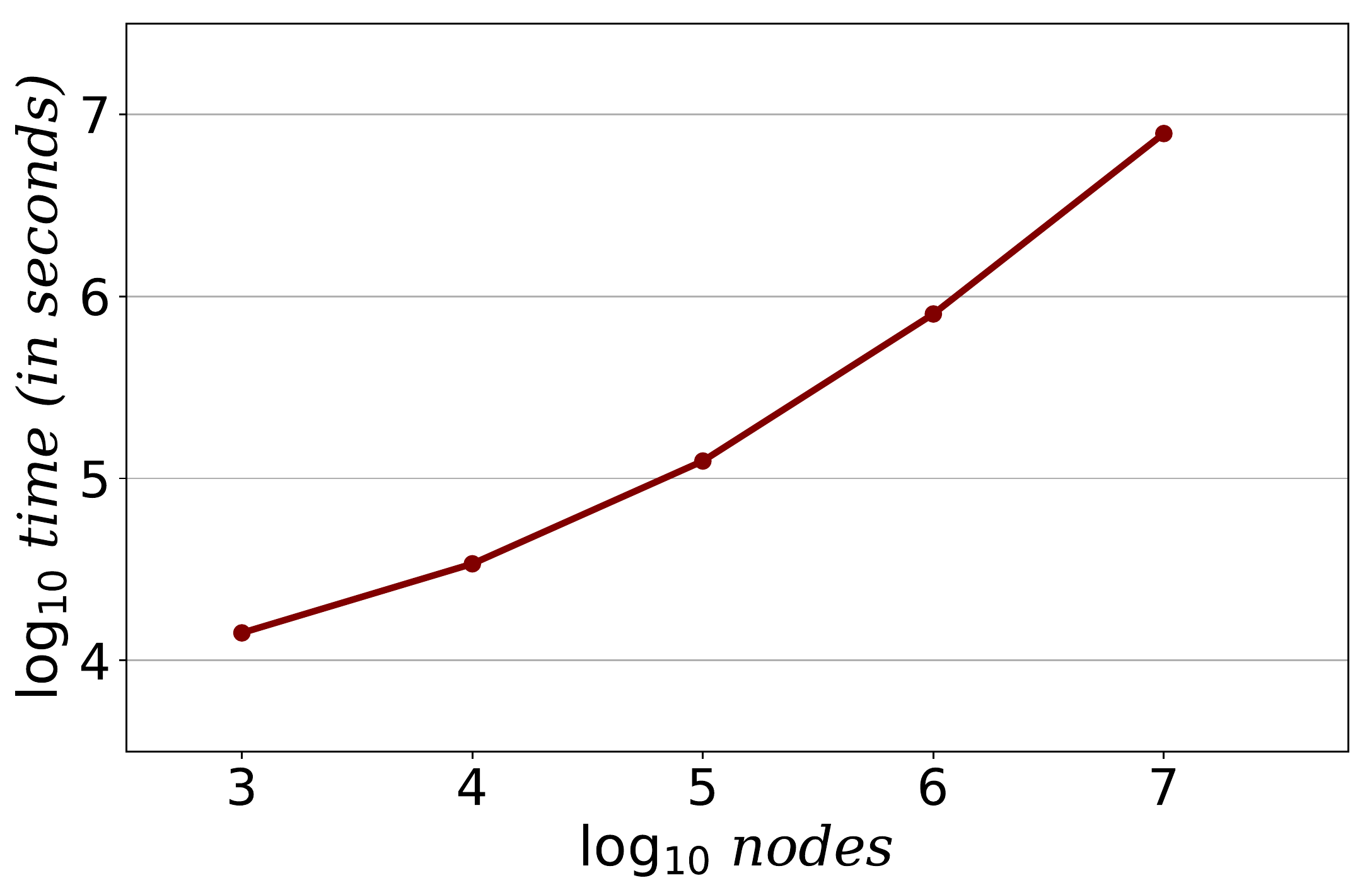}
	\caption{The scalability of \emph{RWNE} over different network sizes.}
	\label{Fig:scalability_time}
% 	\vspace{-0.1in}
\end{figure}

\clearpage

\subsection{Case Study: Scalability}
As analyzed in Section~\ref{sec:comlexity}, the proposed \emph{RWNE} model has superior scalability with linear complexity.
In this section, we further experimentally verify the scalability of \emph{RWNE}.
% \textcolor{blue}{
We first generate a series of random graphs with different sizes of [1k; 10k; 100k; 1000k; 10000k] from the Youtube dataset by randomly choosing several nodes and expanding from them to the fixed size while avoiding isolated nodes, and then apply our method to these synthetic networks to learn node embeddings.
The time consumption is shown in Figure~\ref{Fig:scalability_time}.
We can see that the running time grows linearly with the number of nodes, comparable to the random-walk-based methods (e.g., DeepWalk and node2vec). In the case where the size of the graph is larger, the proposed model has obvious advantages over the methods with higher computational complexity (e.g., GCN and GAT). Thus, our method is efficient and scalable for large-scale networks.
% }

%\vspace{-0.05in}
\section{Conclusions}\label{sec:conclusions}
%\vspace{-0.05in}
In this paper, we present a general scalable random-walk-based network embedding framework \textbf{\emph{RWNE}} to effectively and efficiently preserve higher-order proximity.

Distinguishing from existing random-walk-based methods,
we focus more on an explicit framework to directly leverage random walk to preserve higher-order proximity with carefully designed objective instead of the current two-step approach.
% we systematically design a sound objective to explicitly preserve arbitrary higher-order proximity and then equivalently optimize it by random walk simulation.
We first systematically design a joint objective to simultaneously capture both the local pairwise similarity and the global listwise equivalence provided by arbitrary higher-order proximity.
Then we leverage a node-sampling optimization to equivalently eliminate the computation of higher-order proximity by random walk simulation.

As a result, the random walk is theoretically incorporated into the objective function, which explicitly clarifies the essential role of random walk playing in higher-order proximity preserved network embedding.

Further,
we also introduce the random walk with restart process to naturally and effectively weigh the proximities of different orders by a personalized teleport probability.
%As a result, 
%{\it RWNE} is able to  effectively and efficiently preserve a personalized higher-order proximity and 
%The superiority of our model is demonstrated by the extensive experiments we conducted on node classification and clustering, as well as the parameter analysis.
We conduct extensive experiments on several datasets and the results demonstrate the superiority of our model.

\acks{The authors would like to thank the anonymous reviewers of the Journal of Artificial Intelligence Research (JAIR) for their valuable advice. 
This work was supported by the NSFC through grants (No.U20B2053, 61872022, and 62002007), and the State Key Laboratory of Software Development Environment (SKLSDE-2020ZX-12). 
This work was also sponsored by CAAI-Huawei MindSpore Open Fund. Thanks for computing infrastructure provided by Huawei MindSpore platform. For any correspondence, please refer to Jianxin Li and Hao Peng.
}

% \clearpage
% \clearpage
\appendix

\section{More Results}
\label{appendix:more_results}
% % We report the standard deviations of the Micro-F1 scores and the Macro-F1 scores for multi-label node classification in Table~\ref{Tab:results_Macro_f1} and Table~\ref{Tab:results_Micro_f1}, 
% % the NMI (Normalized Mutual Information) scores for node clustering in Table~\ref{Tab:results_nmi},
% % and the MAP (Mean Average Precision) scores for link reconstruction in Table~\ref{Tab:results_map}.
% The standard deviations for multi-label node classification are shown in Table~\ref{Tab:app_results_Macro_f1} and Table~\ref{Tab:app_results_Micro_f1},
% for node clustering are shown in Table~\ref{Tab:app_results_nmi},
% and for link reconstruction are shown in Table~\ref{Tab:app_results_map}.

As mentioned in Section~\ref{sec:experiments}, we evaluate the quality of the embedding vectors learned by different methods on three classical network mining
tasks: multi-label node classification, node clustering, and link reconstruction.
% To ensure the significance of the results, we repeat each experiment ten times and %use the mean value as the reporting result. 
Here we report their standard deviation values which are absent due to the limitation of the page's width.

%We report the mean values and 
% \textcolor{blue}{
In detail, we report the standard deviations of the Micro-F1 scores and the Macro-F1 scores for multi-label node classification in Table~\ref{Tab:app_results_Macro_f1} and Table~\ref{Tab:app_results_Micro_f1},
the NMI (Normalized Mutual Information) scores for node clustering in Table~\ref{Tab:app_results_nmi},
and the MAP (Mean Average Precision) scores for link reconstruction in Table~\ref{Tab:app_results_map}.

\begin{table}[h]
	\centering
% 	\begin{tabular}{c|c|c|c|c|c|c}
% 		\toprule
% 		Dataset   & Cora   & Pubmed & DBLP & Blogcat. & Flickr & % \begin{tabular}{p{1.7cm}|p{1.6cm}<{\centering}|p{1.6cm}<{\centering}|p{1.6cm}<{\centering}|p{1.6cm}<{\centering}|p{1.6cm}<{\centering}|p{1.6cm}<{\centering}}
\begin{tabular}{p{1.7cm}p{1.6cm}<{\centering}p{1.6cm}<{\centering}p{1.6cm}<{\centering}p{1.6cm}<{\centering}p{1.6cm}<{\centering}p{1.6cm}<{\centering}}
% 	\begin{tabular}{l|c|c|c|c|c|c}
        \toprule
        {\bf Dataset}   & {\bf Cora}   & {\bf Pubmed} & {\bf DBLP}   & {\bf Blogcat.} & {\bf Flickr} & {\bf Youtube}\  \\
		\midrule
		DeepWalk  & 0.0034 & 0.0017 & 0.0029 & 0.0030 & 0.0009 & 0.0013 \\
		LINE      & 0.0098 & 0.0030 & 0.0052 & 0.0012 & 0.0078 & 0.0058 \\
		node2vec  & 0.0041 & 0.0016 & 0.0039 & 0.0023 & 0.0011 & 0.0018 \\
		\midrule
		GraRep    & 0.0028 & 0.0012 & 0.0018 & 0.0013 & 0.0008 & -- \\
		HOPE      & 0.0041 & 0.0022 & 0.0054 & 0.0014 & 0.0009 & -- \\
		\midrule
		SDNE      & 0.0203 & 0.0166 & 0.0164 & 0.0023 & 0.0017 & -- \\
		GCN       & 0.0125 & 0.0073 & 0.0057 & 0.0028 & 0.0026 & -- \\
		GAT       & 0.0183 & 0.0100 & 0.0087 & 0.0031 & 0.0044 & -- \\
		\midrule
		RWNE      & 0.0038 & 0.0023 & 0.0031 & 0.0025 & 0.0012 & 0.0014 \\
		\bottomrule
	\end{tabular}
	\caption{The standard deviation of Macro-F1 scores for multi-label node classification.}
	\label{Tab:app_results_Macro_f1}
\end{table}

\begin{table}[h]
	\centering
% 	\begin{tabular}{c|c|c|c|c|c|c}
% 		\toprule
% 		Dataset   & Cora   & Pubmed & DBLP & Blogcat. & Flickr & Youtube \\
% \begin{tabular}{p{1.7cm}|p{1.6cm}<{\centering}|p{1.6cm}<{\centering}|p{1.6cm}<{\centering}|p{1.6cm}<{\centering}|p{1.6cm}<{\centering}|p{1.6cm}<{\centering}}
\begin{tabular}{p{1.7cm}p{1.6cm}<{\centering}p{1.6cm}<{\centering}p{1.6cm}<{\centering}p{1.6cm}<{\centering}p{1.6cm}<{\centering}p{1.6cm}<{\centering}}
% 	\begin{tabular}{l|c|c|c|c|c|c}
        \toprule
        {\bf Dataset}   & {\bf Cora}   & {\bf Pubmed} & {\bf DBLP}   & {\bf Blogcat.} & {\bf Flickr} & {\bf Youtube}\  \\
		\midrule
		Deepwalk  & 0.0027 & 0.0017 & 0.0023 & 0.0019 & 0.0005 & 0.0009 \\
		LINE      & 0.0065 & 0.0036 & 0.0066 & 0.0014 & 0.0018 & 0.0036 \\
		node2vec  & 0.0035 & 0.0019 & 0.0034 & 0.0026 & 0.0015 & 0.0011 \\
		\midrule
		GraRep    & 0.0028 & 0.0011 & 0.0019 & 0.0014 & 0.0003 & -- \\
		HOPE      & 0.0034 & 0.0025 & 0.0026 & 0.0010 & 0.0006 & -- \\
		\midrule
		SDNE      & 0.0169 & 0.0148 & 0.0050 & 0.0048 & 0.0016 & -- \\
		GCN       & 0.0120 & 0.0036 & 0.0082 & 0.0058 & 0.0021 & -- \\
		GAT       & 0.0162 & 0.0082 & 0.0095 & 0.0061 & 0.0028 & -- \\
		\midrule
		RWNE      & 0.0032 & 0.0017 & 0.0027 & 0.0016 & 0.0009 & 0.0008 \\
		\bottomrule
	\end{tabular}
	\caption{The standard deviation of Micro-F1 scores for multi-label node classification.}
	\label{Tab:app_results_Micro_f1}
\end{table}

\clearpage

\begin{table}[t]
	\centering
% 	\begin{tabular}{c|c|c|c|c|c|c}
% 		\toprule
% 		Dataset   & Cora   & Pubmed & DBLP & Blogcat. & Flickr & Youtube \\
% \begin{tabular}{p{1.7cm}|p{1.6cm}<{\centering}|p{1.6cm}<{\centering}|p{1.6cm}<{\centering}|p{1.6cm}<{\centering}|p{1.6cm}<{\centering}|p{1.6cm}<{\centering}}
\begin{tabular}{p{1.7cm}p{1.6cm}<{\centering}p{1.6cm}<{\centering}p{1.6cm}<{\centering}p{1.6cm}<{\centering}p{1.6cm}<{\centering}p{1.6cm}<{\centering}}
% 	\begin{tabular}{l|c|c|c|c|c|c}
        \toprule
        {\bf Dataset}   & {\bf Cora}   & {\bf Pubmed} & {\bf DBLP}   & {\bf Blogcat.} & {\bf Flickr} & {\bf Youtube}\  \\
		\midrule
		Deepwalk  & 0.0041 & 0.0053 & 0.0031 & 0.0014 & 0.0013 & 0.0006 \\
		LINE      & 0.0089 & 0.0127 & 0.0118 & 0.0014 & 0.0026 & 0.0017 \\
		node2vec  & 0.0049 & 0.0064 & 0.0052 & 0.0015 & 0.0014 & 0.0007 \\
		\midrule
		GraRep    & 0.0090 & 0.0002 & 0.0007 & 0.0014 & 0.0005 & -- \\
		HOPE      & 0.0136 & 0.0005 & 0.0024 & 0.0024 & 0.0006 & -- \\
		\midrule
		SDNE      & 0.0184 & 0.0143 & 0.0103 & 0.0047 & 0.0024 & -- \\
		GCN       & 0.0103 & 0.0054 & 0.0041 & 0.0102 & 0.0067 & -- \\
		GAT       & 0.0180 & 0.0088 & 0.0083 & 0.0133 & 0.0112 & -- \\
		\midrule
		RWNE      & 0.0044 & 0.0062 & 0.0038 & 0.0015 & 0.0015 & 0.0008 \\
		\bottomrule
	\end{tabular}
	\caption{The standard deviation of NMI scores for node clustering.}	
	\label{Tab:app_results_nmi}
\end{table}

\begin{table}
	\centering
% 	\begin{tabular}{c|c|c|c|c|c|c}
% 		\toprule
% 		Dataset   & Cora   & Pubmed & DBLP & Blogcat. & Flickr & Youtube \\
% \begin{tabular}{p{1.7cm}|p{1.6cm}<{\centering}|p{1.6cm}<{\centering}|p{1.6cm}<{\centering}|p{1.6cm}<{\centering}|p{1.6cm}<{\centering}|p{1.6cm}<{\centering}}
\begin{tabular}{p{1.7cm}p{1.6cm}<{\centering}p{1.6cm}<{\centering}p{1.6cm}<{\centering}p{1.6cm}<{\centering}p{1.6cm}<{\centering}p{1.6cm}<{\centering}}
% 	\begin{tabular}{l|c|c|c|c|c|c}
        \toprule
        {\bf Dataset}   & {\bf Cora}   & {\bf Pubmed} & {\bf DBLP}   & {\bf Blogcat.} & {\bf Flickr} & {\bf Youtube}\  \\
		\midrule
		Deepwalk  & 0.0021 & 0.0015 & 0.0009 & 0.0008 & 0.0064 & 0.0080 \\
		LINE      & 0.0105 & 0.0046 & 0.0035 & 0.0028 & 0.0079 & 0.0085 \\
		node2vec  & 0.0030 & 0.0021 & 0.0011 & 0.0009 & 0.0078 & 0.0093 \\
		\midrule
		GraRep    & 0.0006 & 0.0009 & 0.0012 & 0.0009 & 0.0028 & -- \\
		HOPE      & 0.0018 & 0.0015 & 0.0015 & 0.0015 & 0.0031 & -- \\
		\midrule
		SDNE      & 0.0117 & 0.0114 & 0.0157 & 0.0036 & 0.0088 & -- \\
		GCN       & 0.0135 & 0.0120 & 0.0159 & 0.0091 & 0.0056 & -- \\
		GAT       & 0.0130 & 0.0152 & 0.0209 & 0.0124 & 0.0092 & -- \\
		\midrule
		RWNE      & 0.0023 & 0.0016 & 0.0021 & 0.0018 & 0.0072 & 0.0078 \\
		\bottomrule
	\end{tabular}
	\caption{The standard deviation of MAP scores for link reconstruction.}	
	\label{Tab:app_results_map}
\end{table}

% \clearpage

\bibliography{ref}
\bibliographystyle{theapa}

\end{document}